\documentclass[runningheads]{llncs}

\usepackage[mobile]{eccv}

\usepackage{eccvabbrv}
\usepackage{graphicx}
\usepackage{booktabs}
\usepackage[accsupp]{axessibility}  % Improves PDF readability for those with disabilities.
\usepackage{multirow}
\usepackage{wrapfig}
\usepackage{hyperref}
\usepackage{orcidlink}

\usepackage{array}    
\usepackage{caption}    
\usepackage[table]{xcolor}

\definecolor{firstplace}{rgb}{0.56, 0.78, 0.58}
\definecolor{secondplace}{rgb}{0.9, 1.0, 0.9} % Light yellow
\definecolor{thirdplace}{rgb}{1.0, 1.0, 0.8} 
\definecolor{cellgreen}{RGB}{0, 180, 139}
\definecolor{cellred}{RGB}{239,99,75}
\definecolor{cellblue}{RGB}{99,113,250}

\def\method{DINO-Tok}

\begin{document}

\title{DINO-Tok: Adapting DINO for Visual Tokenizers} 
\author{Mingkai Jia$^{1,2}$ \quad
  Mingxiao Li$^{2}$ \quad
  Zhijian Shu$^{2,3}$ \quad 
  Anlin Zheng$^{4}$ \\
  Liaoyuan Fan$^{2}$ \quad
  Jiaxin Guo$^{5}$ \quad 
  Tianxing Shi$^{3}$ \quad
  Dongyue Lu$^{2}$ \quad
  Zeming Li$^{1}$ \\ 
  Xiaoyang Guo$^{2}$ \quad
  Xiaojuan Qi$^{4}$ \quad
  Xiao-Xiao Long$^{3}$ \quad
  Qian Zhang$^{2}$ \\
  Ping Tan$^{1}$\thanks{Co-corresponding Author.} \quad
  Wei Yin$^{2}$ \thanks{Co-corresponding Author. Project Leader.}
  }
\institute{$^1$The Hong Kong University of Science and Technology \quad $^2$Horizon Robotics \quad $^3$Nanjing University \quad $^4$Hong Kong University \\ \quad $^5$The Chinese University of Hong Kong}
% \thanks{footnote}
\titlerunning{DINO-Tok}

\makeatletter
\let\@oldmaketitle\@maketitle%
\renewcommand{\@maketitle}{\@oldmaketitle%
 \centering
    \includegraphics[width=1.0\linewidth]{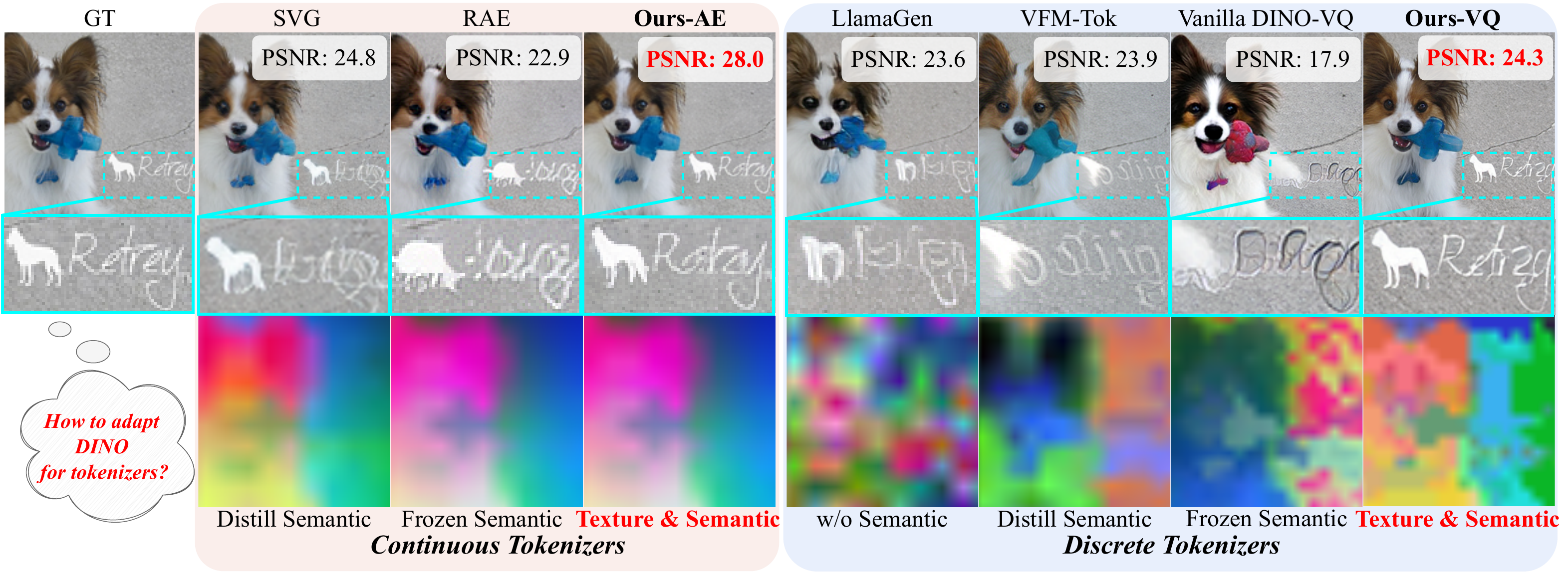}
     \captionof{figure}{
     \small \textbf{How to Adapt DINO for Visual Tokenizers?} \textbf{(Top): RGB reconstructions.} Distilling or freezing DINO features exposes the texture–semantic trade-off such as color shifts and high-dimensional quantization instability or semantic replacement, while \method{}-AE and \method{}-VQ recover fine texture details and correct semantics. \textbf{(Bottom): PCA of latent features.} \method{} produces a more structured semantic latent space, whereas distilled or frozen tokenizers yield noisier ones.}
     \vspace{-25pt}
    \label{fig:teaser}
    \bigskip}            %
\makeatother
\maketitle

\begin{abstract}
Recent advances in visual generation have emphasized the importance of Latent Generative Models (LGMs), which critically depend on effective visual tokenizers to bridge pixels and semantic representations. However, tokenizers constructed on pre-trained vision foundation models (VFMs) often struggle to balance semantic richness and reconstruction fidelity in high-dimensional latent spaces. In this paper, we introduce \textbf{DINO-Tok}, a visual tokenizer built upon a frozen DINO encoder that supports both continuous autoencoding (\textbf{DINO-Tok-AE}) and discrete vector-quantization (\textbf{DINO-Tok-VQ}). By unifying hierarchical representations from both shallow fine-grained features and deep global semantics into an information-complete latent space, DINO-Tok preserves \textit{texture details} while maintaining \textit{semantic consistency} for generation. We further investigate VQ in frozen semantic feature spaces of high dimensionality, where information dilution and codebook collapse frequently arise. To address this issue, we propose \textbf{Dominant-Subspace Quantization (DSQ)}, which leverages a global PCA analysis to select principal components while suppressing noisy dimensions, thereby stabilizing codebook optimization and improving reconstruction and generation quality. On ImageNet 256×256, DINO-Tok achieves strong reconstruction performance, achieving \textbf{0.28} rFID for continuous autoencoding and \textbf{1.10} rFID for discrete VQ, as well as strong few-step generation performance \textbf{1.82} gFID for diffusion and \textbf{2.44} gFID for autoregressive generation. These results demonstrate that pre-trained VFMs such as DINO can be directly adapted into high-fidelity, semantically aligned visual tokenizers for next-generation latent generative models.
\end{abstract}

\section{Introduction}
\label{sec:intro}

Recent advances in large-scale generative models have demonstrated the power of high-level representation learning across vision and language~\cite{yao2025vavae,bai2023qwen,li2023blip}.
In the visual domain, most generative models~\cite{rombach2022ldm, chang2022maskgit} critically rely on visual tokenizers that convert raw pixels into compact latent features.
However, existing tokenizers, including variational autoencoders (VAEs)~\cite{kingma2013vae}, vector quantized variational autoencoders (VQ-VAEs)~\cite{van2017vqvae,esser2021vqgan}, and their recent variants~\cite{yu2023lfq,mentzer2023fsq,tian2025var,chen2024odvae,zhao2024cvvae,zhu2024vqfc,sargent2025flowmo}, are primarily optimized for reconstruction fidelity rather than semantic representation. 
Consequently, the learned latent spaces are often low-dimensional, task-specific, and lack structural alignment with high-level semantics, limiting their expressiveness in modern generative modeling.

\begin{figure}[t]
\begin{center}
\centerline{\includegraphics[width=0.95\textwidth]{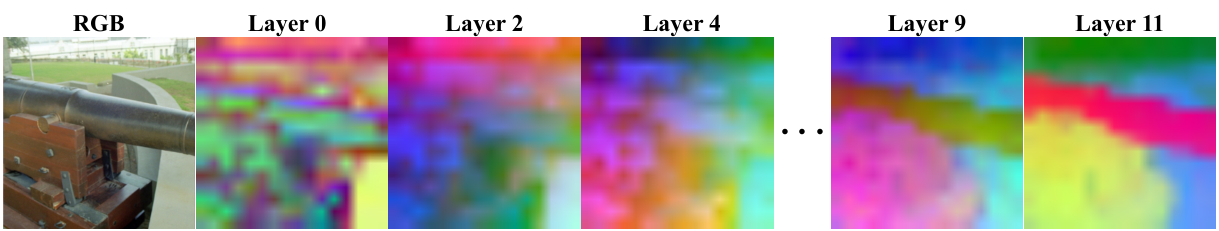}}
\caption{
\textbf{PCA visualizations across shallow to deep layers of DINO-base.} As depth increases, DINO features become more semantically clustered, while fine-grained details fade, indicating increasingly structured representations.
}

\label{fig:dino_layers}
\end{center}
\vspace{-1cm}
\end{figure}

In parallel, pretrained vision foundation models (VFMs) such as DINO~\cite{caron2021dino,oquab2023dinov2,simeoni2025dinov3} have demonstrated strong capabilities in extracting semantically structured and generalizable features. 
Recent works~\cite{yao2025vavae, zheng2025vfmtok} distill VFM~\cite{radford2021clip,oquab2023dinov2,simeoni2025dinov3,tschannen2025siglip} knowledge into tokenizers to improve semantic alignment, while RAE~\cite{zheng2025rae} directly utilizes a frozen VFM as the encoder and observes clear gains in image generation.
However, whether through distillation or direct usage, leveraging VFM features for tokenization still faces two key challenges: (1) \textbf{Texture-semantic trade-off}: Deep VFM layers capture global semantics but the features lack low-level texture information, leading to color shifts, missing fine details, and low PSNR, as depicted in Fig.~\ref{fig:teaser}, \ref{fig:dino_layers}, and \ref{fig:dinov2_dino_rec}; (2) \textbf{High-dimensional quantization instability}: Tokenizers designed for continuous diffusion models cannot be directly converted into discrete tokenizers for autoregressive generation. \textit{e.g.}, directly discretizing high-dimensional frozen VFM features often triggers codebook collapse and semantic replacement, as described in Fig.~\ref{fig:teaser}, \ref{fig:dinov2_dino_rec}. 
These observations motivate a natural question:
\textbf{\textit{Could a pretrained representation model, such as DINO, be directly adapted into an effective unified visual tokenizer for both continuous and discrete generative modeling?}}

To answer this question, we propose \textbf{\method{}}, a representation-driven hybrid tokenizer built on a frozen DINO encoder, comprising a continuous tokenizer \method{}-AE and a discrete tokenizer \method{}-VQ.
To resolve the texture-semantic trade-off, \method{}-AE constructs an \textbf{information-complete latent space} by integrating fine-grained detail features from shallow layers with deep global semantics, preserving high-frequency textures while maintaining semantic consistency. The specified shallow layer could both provide rich high-frequency details to complement semantics and remain sufficiently structured to align with frozen DINO features without additional distillation. Our layer-wise analysis (see Fig.~\ref{fig:ablation_layers}) shows that an intermediate shallow layer best satisfies this trade-off, yielding the most favorable balance between reconstruction quality and generative performance.

For discrete modeling, discretizing such frozen, high-dimensional semantic spaces introduces additional difficulty for VQ. To address this issue, we introduce \textbf{Dominant-Subspace Quantization (DSQ)} in \method{}-VQ. Guided by a global PCA analysis of DINO features, DSQ selects a principal semantic component subspace and performs VQ only along these principal directions while suppressing noisy, low-variance dimensions. This design maintains the effective dimensions to address the quantization problem, restores the discriminative power of distance, and stabilizes codebook optimization, yielding discrete tokens that remain semantically consistent with the underlying frozen VFM. 

This representation-driven design delivers strong empirical performance: \method{} achieves state-of-the-art reconstruction on ImageNet 256$\times$256 ($0.28$~rFID for AE, $1.10$~rFID for VQ) and superior generative quality (few-step $1.82$~gFID for diffusion and $2.44$~gFID for AR) compared to existing methods under identical settings.
Our contributions are threefold:
\begin{itemize}
\item We analyze the challenges of adapting the frozen DINO to vision tokenizers, identifying texture-semantic trade-off and quantization instability in high-dimensional semantic spaces.
\item We propose \textbf{\method{}-AE}, a continuous tokenizer that fuses hierarchical DINO features into an \textbf{information-complete latent space}, achieving high-fidelity reconstruction while preserving semantic structure.
\item We introduce \textbf{\method{}-VQ} with \textbf{dominant-subspace quantization}, which quantizes principal dimensions of frozen DINO features, leading to stable, semantically consistent discrete tokens and improved reconstruction and generation performance.
\end{itemize}

\section{Related Works}
\label{sec:formatting}

\noindent \textbf{Continuous Visual Tokenizers.} Continuous visual tokenizers, often based on variational autoencoders (VAEs) ~\cite{kingma2013vae}, map pixel-level inputs into continuous latent spaces, enabling efficient training of latent diffusion models~\cite{rombach2022ldm, esser2024flow, dai2023emu,xie2024showo,ramesh2021dalle,li2025noisear} with high-fidelity reconstruction.

Recent efforts~\cite{yu2024repa,yao2025vavae,zheng2025rae} have focused on improving the semantic structure of the latent space by leveraging pretrained VFMs~\cite{oquab2023dinov2,kirillov2023sam,he2022mae,radford2021clip}. 
REPA~\cite{yu2024repa} aligns DiT middle block features with representations. 
VA-VAE~\cite{yao2025vavae} further introduces semantic supervision 
via latent-space distillation, encouraging disentangled and meaningful representations. 
Although generation quality improves, weak supervision constrains semantic information retention.
RAE~\cite{zheng2025rae} takes a more direct approach by replacing the VAE encoder with a frozen vision backbone, aiming at stronger semantic priors. 
This setup enhances representational understanding but struggles with fine details and color accuracy.
It often produces outputs resembling semantic substitutions rather than faithful reconstructions.
To overcome this texture–semantic trade-off, we propose a dual-branch tokenizer, named \method{} that fuses DINO’s last-layer features for semantic content and early-layer features for structural details. 
This design improves both reconstruction and semantic, leading to a balanced and effective continuous latent space.

\noindent \textbf{Discrete Visual Tokenizers.}
% \subsection{VQ-VAE.}
Discrete visual tokenizers, typically implemented as VQ-VAEs~\cite{esser2021vqgan,zhang2025revq,shi2025ibq,xiong2025gigatok,wu2025alitok,wang2024omnitokenizer,zhao2024bsq}, encode image features into discrete codes by looking up the nearest entries in a learnable codebook. 
The discrete tokens enable the ability for autoregressive approaches with efficient storage and compatibility with vision language models (VLMs) and world models (WMs)~\cite{agarwal2025cosmos,hu2024drivingworld,kong2025survey}. 
However, recent works such as LlamaGen~\cite{sun2024llamagen} and Emu3~\cite{wang2024emu3} observe that high-dimensional latents severely degrade quantization, leading to poor reconstruction quality and underutilized codebooks. 
As a result, these models reduce the bottleneck dimension to $8$ to stabilize training.

While this low-dimensional latent improves quantization behavior, it introduces significant information loss during compression. 
To alleviate this, recent methods~\cite{yu2023lfq,luo2024oplfq,mentzer2023fsq} adopt lookup-free quantization, directly projecting features into fixed discrete codebook vectors without nearest-neighbor search. 
Others~\cite{zheng2025vfmtok,bachmann2025flextok,li2024imagefolder,bai2024fqgan,yu2024titok} follow a different path by distilling visual foundation model features into the discrete latent space. 
Despite this semantic alignment, they both constrain the latent dimensionality (e.g., LFQ~\cite{luo2024oplfq} uses a binary codebook of size $2^{18}$ with a bottleneck dimension of $18$), which limits representational capacity.
Some methods~\cite{ma2025unitok,jia2025mgvq,zhuang2025wetok,lee2022rqvae,tan2025sweettok} address this by splitting the latent vector into multiple low-dimensional subspaces, each quantized separately. 
While effective for maintaining quantization quality, 
such low per-channel dimensions limit the utilization of high-dimensional semantic information.

In contrast, \method{} starts from the full $768$-dimensional latent space of DINO’s final layer and applies DSQ, which leverages a global PCA analysis to select a principal semantic latent subspace and performs quantization only along these directions, enabling stable and meaningful codebook learning, which overcomes the inherent difficulty of optimizing codebooks in frozen high-dimensional semantic spaces while preserving fine-grained, structured visual information.

\begin{figure*}[t]
\begin{center}
\centerline{\includegraphics[width=1.0\textwidth]{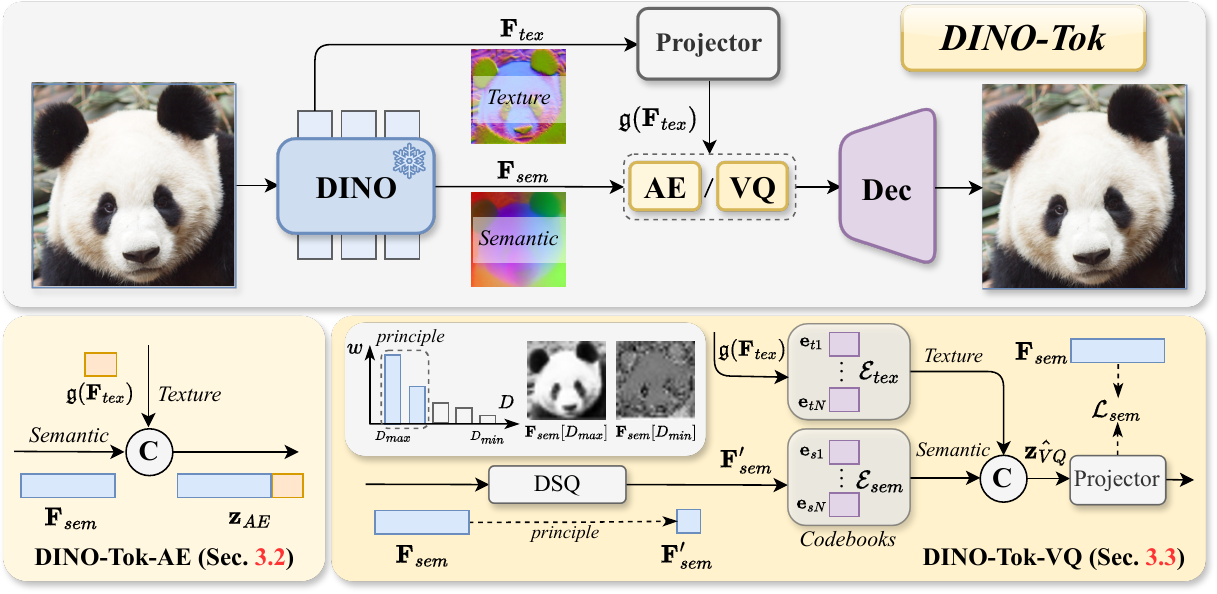}}
\caption{
\textbf{\method{} framework.} We adapt a frozen DINO encoder into a unified visual tokenizer for both continuous and discrete generative modeling. \textbf{\method{}-AE} forms an \textbf{information-complete latent} by concatenating deep semantics with a projected shallow feature to improve reconstruction fidelity. \textbf{\method{}-VQ} discretizes this representation with \textbf{DSQ} and two specialized codebooks for semantics and texture to stabilize high-dimensional lookup while preserving fine details. We further apply an additional semantic alignment loss to retain high-level semantics while maintaining reconstructive quality.
}
\vspace{-1.1cm}
\label{fig:pipeline}
\end{center}
\end{figure*}

\section{DINO-Tok: Adapting DINO for Unified Tokenization}

Recent works~\cite{yao2025vavae,zheng2025vfmtok,ma2025unitok} have demonstrated that incorporating pretrained VFMs into visual tokenizers, either via distillation~\cite{zheng2025vfmtok,shi2025svg} or directly using them as frozen encoders~\cite{zheng2025rae}, can accelerate convergence and improve generative quality.
However, distillation typically transfers only a portion of the semantic structure, while directly using frozen VFM features exposes two fundamental challenges:
\begin{itemize}
    \item \textbf{Texture–semantic trade-off.} Deep VFM features are highly semantic, but lack detailed information such as color and textures that are crucial for high-fidelity reconstruction. 
    \item \textbf{Quantization instability in frozen high-dimensional semantic spaces.} Directly applying VQ to frozen high-dimensional features (e.g., $d=768$ for DINO-Base) suffers from distance measurement failure and semantic imbalance across channels, leading to codebook collapse and semantic replacement.
\end{itemize}
These two challenges lead to the central problem posed in the introduction: \textit{could the frozen representation models be encoders for both continuous and discrete tokenizers}? 
To this end, we design \textbf{\method{}}, a representation-driven hybrid tokenizer built on a frozen DINO~\cite{simeoni2025dinov3} encoder. It comprises a continuous tokenizer, \textbf{\method{}-AE}, which constructs an information-complete latent by fusing shallow texture features with deep semantic features to resolve the texture–semantic trade-off, and a discrete tokenizer, \textbf{\method{}-VQ}, which applies DSQ to stabilize VQ in  frozen high-dimensional semantic spaces by quantizing principle components while preserving critical semantic information.

The overview is illustrated in Fig.~\ref{fig:pipeline}. \method{}-AE and \method{}-VQ shared the same backbone, and we first train AE, then discretize the same fused latent using DSQ.
The method section is organized as follows: Sec.~\ref{sec:challenge} analyzes the detailed challenges of frozen DINO tokenization. Sec.~\ref{sec:dinotok_ae} describes how we design our fused multi-layer dual-branch autoencoder upon frozen VFM DINOv3~\cite{simeoni2025dinov3}. Sec.~\ref{sec:dinotok_vq} shows the DSQ design and dual codebooks following our AE. Sec.~\ref{sec:generation} presents our generative model design. 

\begin{figure}[t]{
\centerline{\includegraphics[width=0.98\textwidth]{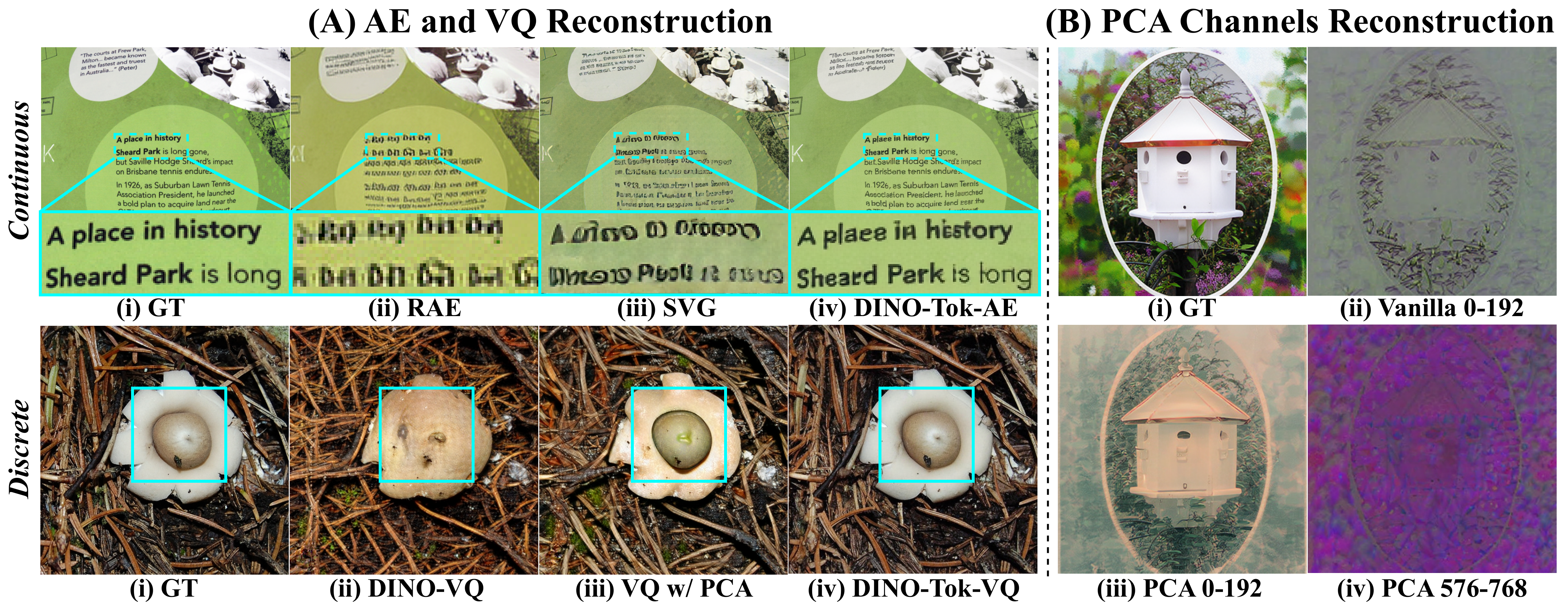}}
\caption{
\textbf{(Left): DINO-based AE/VQ reconstructions.} 
\emph{Top: continuous.} \method{}-AE restores textures and colors via dual-branch features. 
\emph{Bottom: discrete.} Vanilla VQ suffers from semantic replacement and overlap, our dual-branch design with PCA-based VQ achieves faithful reconstruction.
\textbf{(Right): Channel ablation on DINO.} Top-192 PCA-ranked channels (iii) preserve object semantics and structure better than the first 192 raw channels (ii), while the lowest-192 channels (iv) collapse to blurry noise, showing that informative content concentrates in high-eigenvalue components.
}
\vspace{-4mm}
\label{fig:dinov2_dino_rec}}
\end{figure}

\subsection{Challenges of Frozen DINO Tokenization}
\label{sec:challenge}

\noindent\textbf{Texture–semantic trade-off.}
A straightforward baseline, as in RAE~\cite{zheng2025rae}, is to pair a trainable decoder with a frozen DINO encoder and decode directly from the final-layer feature $F_{sem}$. However, Fig.~\ref{fig:teaser} and ~\ref{fig:dinov2_dino_rec} show that reconstructions suffer from color shifts, missing high-frequency textures, and low PSNR. This degradation indicates that $F_{sem}$ is \textit{information-incomplete}: DINO is optimized for semantic structural latent space, suppressing low-level appearance variations that are essential for pixel-level fidelity. Prior work such as SVG~\cite{shi2025svg} partially alleviates this by introducing an extra residual encoder to recover fine details from pixels, but the underlying texture–semantic trade-off remains.

\noindent\textbf{High-Dimensional Quantization Instability.}
Extending a pretrained visual foundation model to build a discrete tokenizer~\cite{van2017vqvae} introduces further complications. 
As dimensionality increases, pairwise distances tend to concentrate~\cite{beyer1999nearest}: the ratio between nearest and farthest neighbor distances approaches $1$, making $L_2$-based nearest-neighbor assignments numerically indistinguishable. See supplementary materials for detailed analysis.
In frozen DINO semantic spaces, this distance concentration leads to eventual codebook collapse (See Tab.~\ref{tab:ab_pca_quant}), together with semantic replacement artifacts (See  Fig.~\ref{fig:dinov2_dino_rec}) in the reconstructed images. 
Meanwhile, a global PCA analysis on DINO features (Fig.~\ref{fig:pca_compare}) and vanilla DINO PCA channels reconstructions (Fig.~\ref{fig:dinov2_dino_rec}-(B)) reveal a pronounced semantic imbalance across channels: a small set of principal components captures coherent semantic and spatial structure, while many tail dimensions are low-variance and noisy. 
Uniform quantization over all channels therefore dilutes dominant semantic components and wastes capacity on noisy dimensions, further aggravating quantization instability in frozen high-dimensional semantic spaces.
Existing approaches~\cite{sun2024llamagen,wang2024emu3} typically utilize very low-dimensional codebooks (e.g., $8$ channels) to mitigate codebook collapse, but such aggressive compression sacrifices latent information capacity. 
High-dimensional VQ methods~\cite{shi2025ibq,zhu2025simvq} rely on jointly optimizing all codes simultaneously, and thus are hard to align with frozen encoded semantic features, whose structural latent space we aim to preserve. These limitations motivate our \textit{dominant-subspace} view of quantization.

\subsection{DINO-Tok-AE: Information-complete Latents}
\label{sec:dinotok_ae}

To resolve the \textbf{texture–semantic trade-off} of the frozen DINO features, \method{}-AE constructs an \textit{information-complete} dual-branch latent that explicitly fuses deep semantic features with shallow texture features which are necessary for faithful reconstruction. (See Fig.~\ref{fig:dino_layers}.)

Let a frozen DINO encoder extract hierarchical features $\{\mathbf{F}_l\}_{l=1}^{L}$ from an input image $\mathbf{x}\in\mathbb{R}^{H\times W\times 3}$, where $\mathbf{F}_l\in\mathbb{R}^{H_l\times W_l\times C_l}$ is the feature map at layer $l$. Denote the final-layer feature as $\mathbf{F}_{sem}$ and a selected shallow-layer feature as $\mathbf{F}_{tex}$. 
We first project the texture feature into a compact embedding using a lightweight projector $\mathfrak{g}(\cdot)$, and then build the \textbf{information-complete latent} by channel-wise concatenation: 
$\mathbf{z}_{\text{AE}} = [\mathbf{F}_{sem} \,;\, \mathfrak{g}(\mathbf{F}_{tex})]$,
where $[\cdot;\cdot]$ denotes concatenation. Intuitively, $\mathbf{F}_{sem}$ anchors global semantics and layout, while $\mathfrak{g}(\mathbf{F}_{tex})$ injects fine-grained texture details that are suppressed by semantic latents.

We decode $\mathbf{z}_{\text{AE}}$ with a trainable decoder $D_{\text{AE}}$ to reconstruct the image:
$\hat{\mathbf{x}}_{\text{AE}} = D(\mathbf{z}_{\text{AE}})$, and optimize $D_{\text{AE}}$ with traditional reconstruction loss $\mathcal{L}_{rec}$, perceptual loss $\mathcal{L}_{perc}$ and GAN loss $\mathcal{L}_{GAN}$ following VQGAN~\cite{esser2021vqgan}, denoted as:
\begin{equation}
\mathcal{L}_\text{AE}=\lambda_{rec}\mathcal{L}_{rec}+\lambda_{perc}\mathcal{L}_{perc}+\lambda_{GAN}\mathcal{L}_{GAN},
\end{equation}
and the DINO~\cite{simeoni2025dinov3} encoder is kept frozen. The dual-branch design reduces inversion difficulty by enabling direct use of explicit texture cues from $\mathbf{F}_{tex}$, rather than recovering fine-grained textures from semantic features alone.
The resulting continuous latent $\mathbf{z}_{\text{AE}}$ thus becomes both semantically structured and texture-complete, forming a strong basis for subsequent discretization.

\subsection{DINO-Tok-VQ: Dominant-Subspace Quantization}
\label{sec:dinotok_vq}

We build \method{}-VQ by discretizing the fused dual-branch latent while respecting the structure of frozen DINO features. Since direct VQ on the high-dimensional space with $L_2$ distance is challenging due to distance concentration and semantic imbalance, we propose \textbf{DSQ}, which performs VQ in a principal semantic latent subspace rather than in the full high-dimensional space.

\begin{figure}[t]
\begin{center}
\centerline{\includegraphics[width=0.98\textwidth]{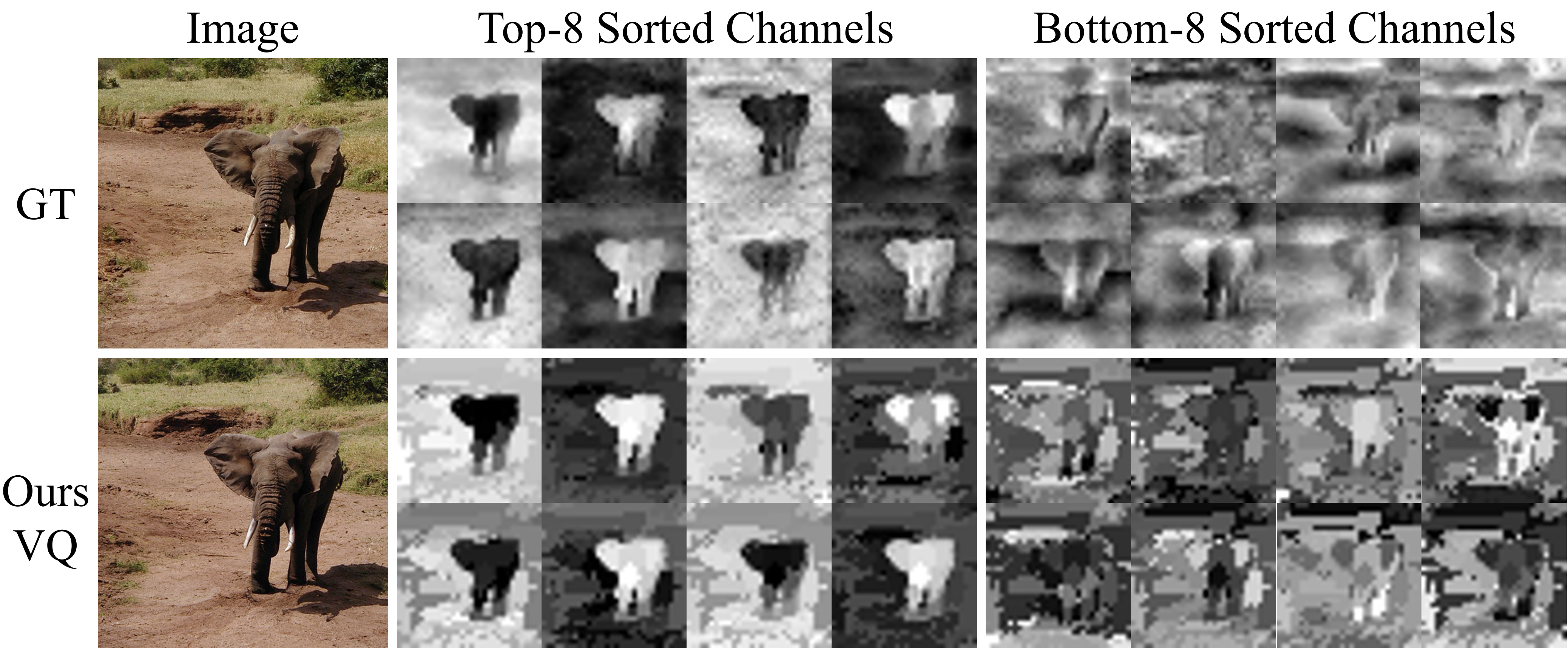}}
\vspace{-3pt}
\caption{
\textbf{Visualization of PCA-sorted feature channels.}
\textbf{PCA-sorted channel visualization.} Channels are ordered by global PCA eigenvalues. High-ranked channels capture clear spatial structures, while low-ranked ones are noisy. Our PCA-guided VQ stabilizes quantization and improves reconstruction fidelity.}
\label{fig:pca_compare}
\vspace{-10mm}
\end{center}
\end{figure}

\noindent\textbf{Global PCA on frozen DINO features.}
We first perform a global PCA on $\mathbf{F}_{{sem}}$ computed over a large ImageNet subset offline. Let $\mathbf{P} \in \mathbb{R}^{C_{{sem}} \times k}$ be the projection matrix formed by the top-$k$ eigenvectors ($k \ll C_{{sem}}$). For each token $i$, we obtain the projected semantic feature ${\mathbf{F}_{{sem}}'}^{(i)}$.
This projection selects a \textit{principle semantic latent subspace} where most semantic structure is concentrated in these dimensions, while many noisy, low-variance channels are discarded. (See Fig.~\ref{fig:pca_compare}.) Reducing dimensionality in this way mitigates distance concentration for VQ while preserving critical semantic information.

\noindent\textbf{Dual-branch codebooks.}
Attaining the dual branch design of \method{}-AE, we maintain two codebooks, including a semantic codebook $\mathcal{E}_{{sem}} = {\mathbf{e}_{{s}k}}$, ${k=1,\dots,N}$ operating in the dominant semantic subspace, and a texture codebook $\mathcal{E}_{{tex}} = {\mathbf{e}_{{t}k}}$, ${k=1,\dots,N}$ operating on the projected texture embeddings with codebook size $N$. For each token $i$, we quantize by:
\begin{align}
\mathbf{q}_{sem}^{(i)} &= \arg\min_{\mathbf{e}_{sk} \in \mathcal{E}_{sem}} || (\mathbf{F}_{sem}'^{(i)} - \mathbf{e_{sk}} ) ||_2^2, \quad \mathbf{q_{tex}^{(i)}} = \arg\min_{e_{tk} \in \mathcal{E}_{tex}} || \mathbf{F}_{tex}^{(i)} - \mathbf{e_{tk}} ||_2^2.
\end{align}
The resulting quantized latent at location $i$ is then: $\hat{\mathbf{z}}_{\text{VQ}}^{(i)} = [\mathbf{q}_{sem}^{(i)}, \mathbf{q}_{tex}^{(i)}],$
and the full latent $\hat{\mathbf{z}}_{\text{VQ}}$ is projected back to the high dimension by $\mathfrak{g_\theta(\hat{\mathbf{z}}_{\text{VQ}})}$ and then fed to the decoder $D$ to obtain the reconstructed image $\hat{\mathbf{x}}_{\text{VQ}} = D(\mathfrak{g_\theta(\hat{\mathbf{z}}_{\text{VQ}})})$. 
\noindent\textbf{Semantic consistency regularization.}
To ensure that discretization does not distort the frozen semantic structure, we add a semantic alignment loss. A small projector $\mathfrak{g}_{\theta}(\cdot)$ maps $\hat{\mathbf{z}}_{\text{VQ}}$ back to the original high-dimensional semantic space, and we penalize deviation from $\mathbf{F}_{\text{sem}}$ via cosine similarity by $\mathcal{L}_{sem}=1-\text{cos}(\mathfrak{g}_\theta(\hat{\mathbf{z}}_{\text{VQ}}),\mathbf{F}_{sem})$.
This term regularizes token alignment with the frozen DINO semantics, enhancing semantic structure under DSQ.

\noindent\textbf{Training objective.}
Following VQGAN~\cite{esser2021vqgan}, we use codebook and commitment losses $\mathcal{L}_{\text{VQ}}$ in addition to $\mathcal{L}_{\text{AE}}$ and $\mathcal{L}_{sem}$:
\begin{align}
% \begin{aligned}
% \end{aligned}
\mathcal{L}
=\mathcal{L}_{AE}+\mathcal{L}_{VQ}+\mathcal{L}_{sem}.
\end{align}
We keep the DINO encoder frozen and optimize codebooks, projector heads, and the decoder. DSQ thus stabilizes VQ in frozen high-dimensional semantic spaces by (i) quantizing in a principle semantic latent subspace and (ii) enforcing semantic consistency to the original DINO features.

\begin{table*}[t]
\centering
\caption{
% \textbf{ ImageNet-1k 256$\times$256 class-to-image generation evaluation.} \method{}-AE-XL with much larger latent dimensions attains a remarkable performance with only $80$ epochs training and $25$ steps. And both \method{}-AE and \method{}-VQ shows scalability for better performance with more training epochs.
\textbf{Comparative evaluation of \method{}-AE on ImageNet-1k 256$\times$256 reconstruction and class-conditional generation.} \method{}-AE is trained only on ImageNet-1k and achieves notable reconstruction performance. For generation, \method{}-AE-XL benefits from larger latent dimensions, reaching competitive performance with only 80 training epochs and 25 sampling steps. $^{*}$ denotes results without classifier-free guidance (CFG).
}
\label{tab:imagenet256ae}
\vspace{-5pt}
\resizebox{0.98\textwidth}{!}{
\setlength{\tabcolsep}{1.5mm}
\begin{tabular}{l|cc|ccccccccc}
\toprule
\multirow[c|]{2}{*}{\textbf{Method}} &
\multicolumn{2}{c|}{\cellcolor{cellred!10}\textbf{Reconstruction}} &
\multicolumn{6}{c}{\cellcolor{cellblue!10}\textbf{Generation}} \\
& {Tokenizer}&   {rFID}$\downarrow$ & {\#Epoch}  & {\#Step} &
{gFID}*$\downarrow$ & {IS}*$\uparrow$ &
{gFID}$\downarrow$ & {IS}$\uparrow$ \\
\midrule
MaskDiT-XL~\cite{zheng2023maskdit} & SD-VAE& 0.87& 1600 &250& 5.69 & 177.9  & 2.28 & 276.6  \\
DiT-XL~\cite{peebles2023ditxl} & SD-VAE & 0.87 &1600 &250&  9.62 & 121.5  & 2.27 & 278.2 \\
SiT-XL~\cite{ma2024sit} &   SD-VAE & 0.87& 1600 &250& 8.61 & 131.7  & 2.06 & 270.3 \\
Faster-DiT~\cite{yao2024fasterdit} & SD-VAE& 0.87 & 400 & 250 &7.91 & 131.3 &  2.03 & 264.0 \\
MDT~\cite{gao2023mdt}&SD-VAE& 0.87&1300&250 &6.23 &143.0  & 1.79 & 283.0 \\
REPA-XL~\cite{yu2024repa} & SD-VAE & 0.87 & 800 &250&  5.90  & -- & 1.42 & 305.7 \\
SiT-XL~\cite{yu2024repa} & VA-VAE& 0.28 & 800 & 250 & 5.96 & 128.0  & 3.63 & 290.6\\
\midrule
SiT-XL~\cite{yu2024repa} &  VA-VAE & $\mathbf{0.28}$ & 80 &25& 7.29 & 121.0 & 4.13 & $\mathbf{279.7}$ \\
SVG-XL~\cite{yu2024repa} &  SVGTok & 0.65& 80 &25& 6.57 & 137.9& \underline{3.54} & 207.6\\
RAE-XL~\cite{zheng2025rae} &  RAE & \underline{0.49} & 80 & 25 & $\mathbf{2.32}$ & $\mathbf{204.8}$  & -- & -- \\
\rowcolor{cellgreen!10}
\textbf{Ours-AE-XL} & \method{}-AE & $\mathbf{0.28}$ & 80 &  25 &  \underline{3.03} & \underline{176.7}  & $\mathbf{2.27}$ & \underline{221.8}\\
\midrule

SVG-XL~\cite{yu2024repa} &  SVGTok & 0.65& 500 &25& 3.94 & 169.3  & 2.10 & 258.7\\
SVG-XL~\cite{yu2024repa} &  SVGTok & 0.65& 1400 &25& 3.36 & 181.2  & \underline{1.92} & \underline{264.9} \\

RAE-XL~\cite{zheng2025rae} &  RAE& \underline{0.49} & 800 &25& $\mathbf{1.76}$ & $\mathbf{237.1}$  & -- & -- \\
\rowcolor{cellgreen!10}
\textbf{Ours-AE-XL} & \method{}-AE& $\mathbf{0.28}$ & 800 &  25 & \underline{2.40} & \underline{216.2}  & $\mathbf{1.82}$ & $\mathbf{273.7}$ \\
\arrayrulecolor{black}\bottomrule
\end{tabular}
}
\end{table*}

\begin{table*}[t]
\centering
\caption{
\textbf{Comparative evaluation of \method{}-VQ on ImageNet-1k 256$\times$256 reconstruction and autoregressive class-conditional generation.}
}
\vspace{-5pt}
\label{tab:imagenet_gen_ar}
\resizebox{0.95\textwidth}{!}{
\setlength{\tabcolsep}{3.0mm}
\begin{tabular}{l|cc|ccc}
\toprule

\multirow[c|]{2}{*}{\textbf{Method}} &
\multicolumn{2}{c|}{\cellcolor{cellred!10}\textbf{Reconstruction}} &

\multicolumn{3}{c}{\cellcolor{cellblue!10}\textbf{Generation}} \\
 &  {Tokenizer}&   {rFID}$\downarrow$   & {\#Param.} &
{gFID}$\downarrow$ & {IS}$\uparrow$ 
\\
\midrule
LlamaGen-L~\cite{sun2024llamagen} &VQGAN~\cite{esser2021vqgan} & 4.98  & 343M & 3.81 & 248.3 \\
LlamaGen-L~\cite{sun2024llamagen} & TiTok-L~\cite{yu2024titok} & 2.21  & 343M & 4.03 & 219.5  \\
MaskGIT~\cite{chang2022maskgit}& MaskGiT~\cite{chang2022maskgit} & 2.28  & 675M & 4.02 & $\mathbf{355.6}$ \\
VAR-d16~\cite{tian2025var} & VAR~\cite{tian2025var} & -  & 310M & {3.30} & {274.4}  \\
VFMTok-L~\cite{zheng2025vfmtok} & VFMTok & 1.13  & 343M & \underline{2.75} & \underline{278.8}  \\

\rowcolor{cellgreen!10}
\textbf{Ours-VQ-L} & \method{}-VQ & \textbf{1.10}  & 343M &  $\mathbf{2.66}$ & 247.7 \\
\midrule
VQGAN~\cite{esser2021vqgan} & VQGAN~\cite{esser2021vqgan} & 4.98 & 1.4B & 15.78 & 74.3 \\
LlamaGen-XXL~\cite{sun2024llamagen} &VQGAN~\cite{esser2021vqgan}& 4.98  & 1.4B & \underline{3.08} & \underline{253.6} \\
RQ-Transformer~\cite{lee2022rqvae} & RQ-VAE~\cite{lee2022rqvae} & \underline{3.20} & 1.4B & 8.71 & - \\
\rowcolor{cellgreen!10}
\textbf{Ours-VQ-XXL} & \method{}-VQ & \textbf{1.10}   & 1.4B & $\mathbf{2.44}$ & $\mathbf{254.2}$  \\
\arrayrulecolor{black}\bottomrule
\end{tabular}
}
\end{table*}

\subsection{Image Generation with DINO-Tok}
\label{sec:generation}
To assess the usefulness of \method{} for downstream generation, we plug it into two representative generative paradigms: \textit{continuous-space generation} operating on \textit{continuous} latents and and \textit{discrete generation} operating on \textit{discrete} tokens. Concretely, we follow the generation setups of VAVAE~\cite{yao2025vavae} for diffusion and LlamaGen~\cite{sun2024llamagen} for autoregressive modeling, while replacing their tokenizers with our \method{}-AE and \method{}-VQ, respectively. 

\noindent\textbf{Continuous-space generation with \method{}-AE.}
For continuous-space generation, an image is encoded by our fused multi-layer dual-branch tokenizer into the texture-augmented semantic latent $\mathbf{z}_{\text{AE}} = [\mathbf{F}_{sem} \,;\, \mathfrak{g}(\mathbf{F}_{tex})]$, which is decoded to pixels by $D$. We then train a latent generator $G_{\text{AE}}$ to model the distribution of $\mathbf{z}_{\text{AE}}$.
Since $\mathbf{z}_{\text{AE}}$ consists of a semantic branch and a texture branch, we compute the latent objective separately on the two channel subsets and average them:
$\mathcal{L}_{\text{latent}}
= \tfrac{1}{2}\big(\mathcal{L}_{\text{latent}}^{{sem}} + \mathcal{L}_{\text{latent}}^{{tex}}\big),$
which encourages balanced modeling of both semantics and textures rather than collapsing to a single branch.
At sampling time, $G_{\text{AE}}$ produces a latent $\tilde{\mathbf{z}}_{\text{AE}}$ in the same space, which is rendered into the final image by $D$.

\noindent\textbf{Discrete-token generation (\method{}-VQ).}
For discrete-token generation, each spatial location is mapped to two sub-codes by our dual-codebook quantization, one from the semantic codebook $\mathcal{E}_{{sem}}$ and the other from the texture codebook $\mathcal{E}_{{tex}}$.
We adopt the transformer backbone of LlamaGen~\cite{sun2024llamagen} as a discrete generator $G_{\text{disc}}$ and adapt its output layer to our two-codebook setting. The shared backbone produces a hidden state $h^{(i)}$ at each token position, which is fed into two separate linear heads to predict logits for the semantic and texture codebooks respectively, $\boldsymbol{\ell}_{\text{sem}}^{(i)} = W_{\text{sem}} h^{(i)}, \quad
\boldsymbol{\ell}_{\text{tex}}^{(i)} = W_{\text{tex}} h^{(i)}$.
The training loss is the sum of two cross-entropy terms,
$\mathcal{L}_{\text{disc}}
= \mathcal{L}_{\text{CE}}(\boldsymbol{\ell}_{\text{sem}}, \mathbf{y}_{\text{sem}})
+ \mathcal{L}_{\text{CE}}(\boldsymbol{\ell}_{\text{tex}}, \mathbf{y}_{\text{tex}})$,
where $\mathbf{y}_{\text{sem}}$ and $\mathbf{y}_{\text{tex}}$ are ground-truth semantic and texture code indices produced by \method{}-VQ.
At inference time, $G_{VQ}$ autoregressively generates both sub-codes at each position, which are then mapped back by $\mathcal{E}_{\text{sem}}$ and $\mathcal{E}_{\text{tex}}$, concatenated into $\hat{\mathbf{z}}_{\text{VQ}}$, and finally decoded into pixels by the shared decoder $D$.

\begin{figure*}[thbp]
\begin{center}
\centerline{\includegraphics[width=1.0\textwidth]{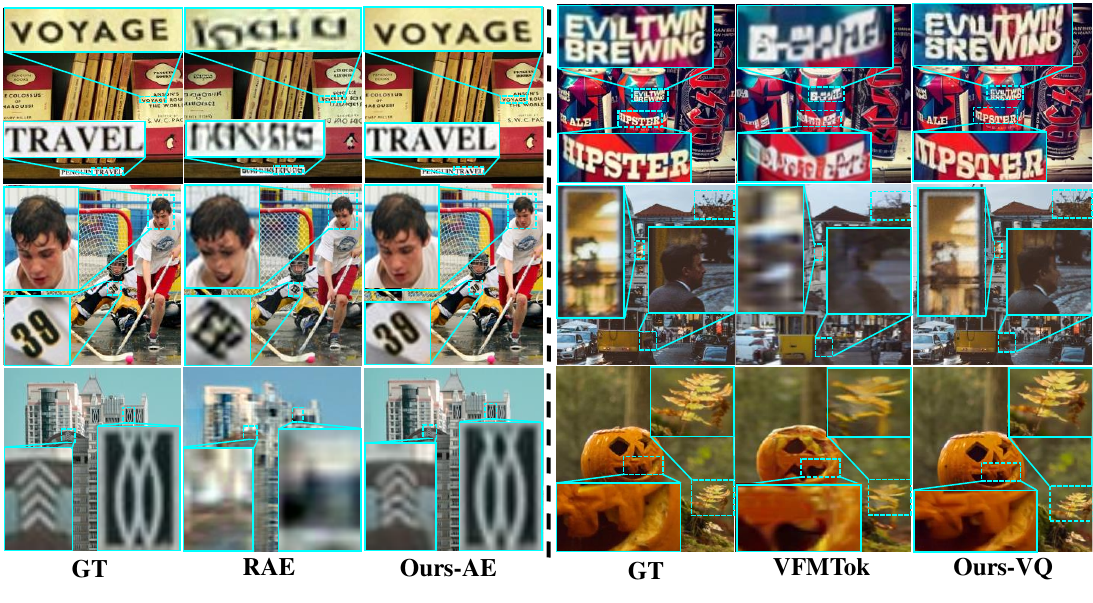}}
\vspace{-.35cm}
\caption{
\textbf{Qualitative reconstruction results under 16× downsampling.} Zoom in for detailed texture comparison. Continuous tokenizers results are shown on the left, and discrete tokenizers on the right. Both \method{}-AE and \method{}-VQ can reconstruct more faithful details and higher-fidelity images compared to the baselines.
}
\vspace{-1.3cm}
\label{fig:rec_qualitative}
\end{center}
\end{figure*}
\section{Experiments}

\label{sec:experiments}

\subsection{Implementation Details}
\noindent\textbf{Training Setup.}
Since the designed visual tokenizers are built upon a frozen pre-trained vision foundation model, only the parameters of newly introduced modules, including the projector, quantizer, and decoder, are learnable. We first train the autoencoders (\method{}-AE),  then introduce discrete quantization and finetune it to the \method{}-VQ. All the training procedures employ a base learning rate of 1e-4. We use the AdamW\cite{loshchilov2017adamW} optimizer with $\beta_1$ and $\beta_2$ as $0.9$ and $0.95$. All models are trained on ImageNet-1k\cite{deng2009imagenet}  training set while evaluated on the validation set with image resolution of $256\times256$. Additionally, we strictly follow VAVAE ~\cite{yao2025vavae} for continuous-space generation and LlamaGen~\cite{sun2024llamagen} for discrete-space generation, respectively.

\noindent\textbf{Evaluation Setting.}
For image reconstruction evaluation, we employ reconstruction Fréchet Inception Distance (rFID) to measure the quality, and for image generations, we employ gFID and inception score (IS) for evaluation, following the standard protocol in prior work~\cite{yao2025vavae, sun2024llamagen, ma2024sit, zheng2025rae}.

\subsection{Comparative Studies}
\noindent\textbf{Reconstruction Quality Comparison.}
Table~\ref{tab:imagenet256ae} and Table~\ref{tab:imagenet_gen_ar} report reconstruction rFID on ImageNet-1k at $256\times256$ for both continuous (\method{}-AE) and discrete (\method{}-VQ) tokenizers. On the continuous side, \method{}-AE-XL attains an rFID of $\mathbf{0.28}$, matching VA-VAE~\cite{yao2025vavae} ($0.28$ rFID) and clearly outperforming representation-based tokenizers such as RAE-XL~\cite{zheng2025rae} ($0.49$ rFID) and SVG-XL~\cite{yu2024repa} ($0.65$ rFID), as well as SD-VAE based frameworks ($0.87$ rFID). This confirms that our dual-branch information-complete latent effectively resolves the texture–semantic trade-off.
On the discrete side, \method{}-VQ reaches an rFID of $\mathbf{1.10}$, significantly improving over VQGAN used in LlamaGen ($4.98$ rFID) and also surpassing other methods, indicating that DSQ preserves critical information for reconstruction under vector quantization.

\noindent\textbf{Continuous-space class-conditional generation.}
We next evaluate class-conditioned generation in continuous-space by plugging \method{}-AE into the continuous generator (Table~\ref{tab:imagenet256ae}).
We focus on the few-step generation with only $25$ sampling steps.
Under this setting and $80$ epochs of training, \method{}-AE-XL achieves a gFID of $\mathbf{2.27}$ with classifier-free guidance (CFG) and $3.03$ without CFG, outperforming SVG-XL~\cite{shi2025svg} ($3.54$ gFID with CFG) and VA-VAE~\cite{yao2025vavae} ($4.13$ gFID with CFG), and approaching RAE-XL~\cite{zheng2025rae} ($2.32$ gFID without CFG), while offering much stronger reconstruction ($0.28$ v.s. $0.49$ rFID) quality and preserving the frozen semantic structure.
When training is extended to $800$ epochs, \method{}-AE-XL further improves to a gFID of $\mathbf{1.82}$ and IS of $\mathbf{273.7}$ with $25$ steps, surpassing SVG-XL~\cite{shi2025svg} at the same step budget and clearly outperforming VA-VAE~\cite{yao2025vavae} in the few-step setting.
These results show that an information-complete latent from \method{}-AE not only preserves reconstruction fidelity, but also enables efficient few-step continuous-space generation with competitive or superior performance compared to prior tokenizers.

\noindent\textbf{Discrete-token class-conditional generation.}
Finally, we assess discrete-token generation by combining \method{}-VQ with the LlamaGen-style autoregressive generator (Table~\ref{tab:imagenet_gen_ar}).
At the L-scale (343M parameters), \method{}-VQ-L attains a gFID of $2.66$ and IS of $247.7$ with rFID $1.10$, improving gFID over LlamaGen-L ($3.81$ gFID, $248.3$ IS) while delivering much better reconstruction and semantic fidelity.
At the XXL scale (1.4B parameters), \method{}-VQ-XXL achieves $2.44$ gFID and $254.2$ IS, outperforming the baseline LlamaGen-XXL ($3.08$ gFID, $253.6$ IS), showing the scaling ability of our \method{}-VQ tokenizer for image generation.
Overall, these results demonstrate that DSQ enables high-dimensional, semantics-preserving VQ that remains compatible with strong autoregressive generators, yielding discrete-token models that jointly excel in reconstruction and class-conditional image generation.

\noindent\textbf{Qualitative results.}
We further visualize reconstruction quality in Fig.~\ref{fig:rec_qualitative}. Compared to previous tokenizers~\cite{esser2024sd3,zheng2025vfmtok}, our approach yields higher fidelity, preserving fine textures and small structures in challenging regions such as faces and embedded text. In contrast, prior methods often produce blurred or distorted content in these areas, or fail to reconstruct them at all, whereas DINO-Tok maintains sharp contours, legible characters, and semantically consistent details.

\begin{figure*}[t]
\begin{center}
\centerline{\includegraphics[width=0.97\textwidth]{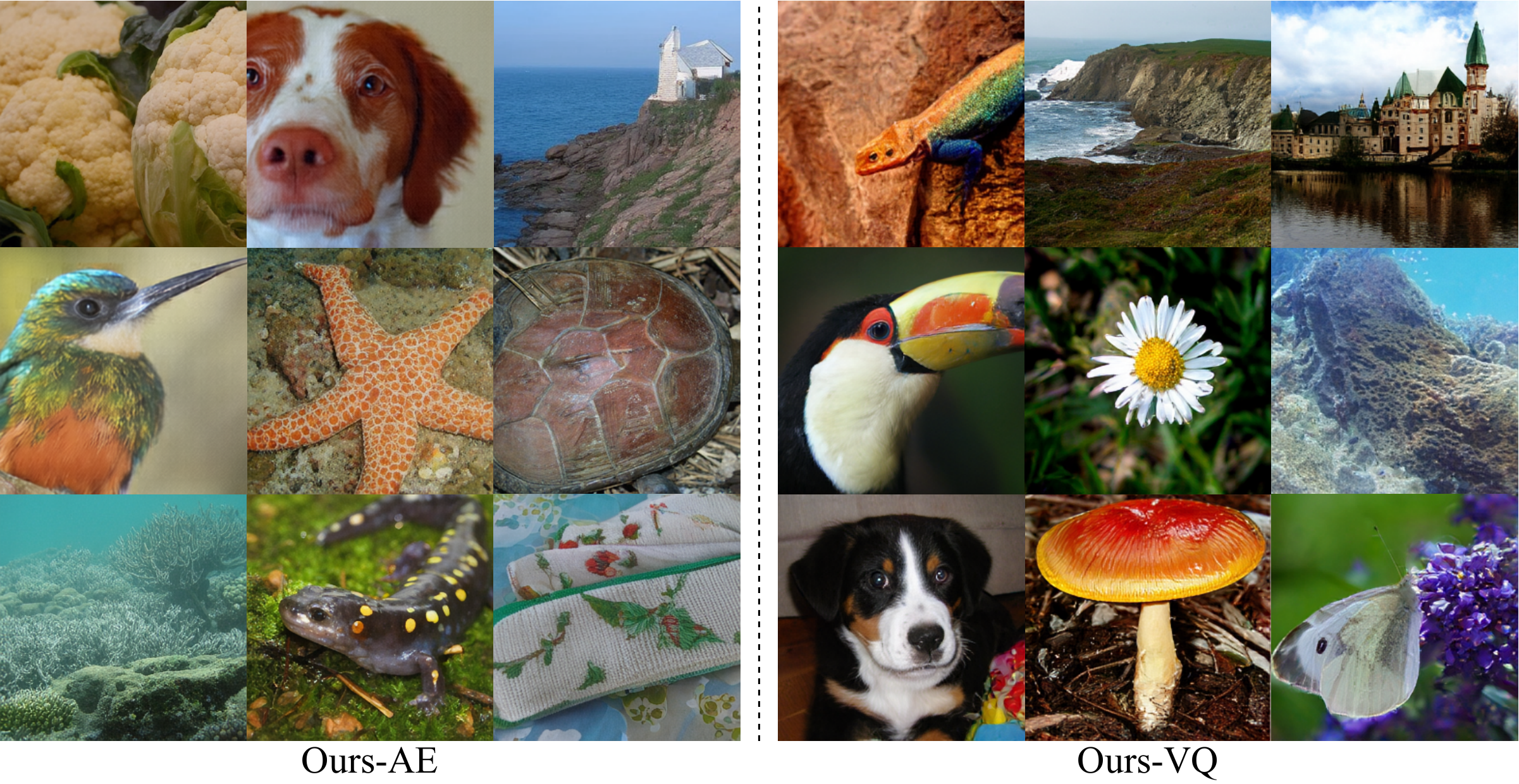}}
\vspace{-5pt}
\caption{
\textbf{Class-conditional image generation examples on ImageNet-1k 256×256.} Left: DINO-Tok-AE (continuous latent generation). Right: DINO-Tok-VQ (discrete-token generation). Both branches produce high-quality samples with fine details and broad diversity across different classes. 
}
\label{fig:gen_qual}
% \vspace{-20pt}
\end{center}
\vspace{-1.cm}
\end{figure*}

In Fig.~\ref{fig:gen_qual}, we show class-conditional samples for both continuous- and discrete-space generation, with \method{}-AE on the left and \method{}-VQ on the right. The \method{}-AE generator produces diverse, high-fidelity images with coherent global structure and rich local textures, reflecting the information-complete latent space built from fused semantic and texture features. The \method{}-VQ generator, despite operating on discrete tokens, preserves most of these semantics and fine details, avoiding typical VQ artifacts such as semantic replacement or overly smoothed textures. Across a wide range of classes, both branches exhibit sharp object boundaries, consistent colors, and recognizable fine-grained attributes, confirming that DINO-Tok provides a robust latent interface for both continuous and discrete generative models.

\subsection{Ablation Study}
We conduct a comprehensive ablation study on \method{} related to texture feature selection and dominant-subspace quantization. More ablations and analyses are presented in the supplement.
% We conduct a comprehensive ablation study on \method{}. 
% More ablations and analyses appear in the supplement.

\begin{figure*}[t]
\centering
\begin{minipage}{.48\textwidth}
\centering
\label{tab:ab_dec_size}
\resizebox{1.0\linewidth}{!}{
\setlength{\tabcolsep}{1.8mm}
\begin{tabular}{l|ccccc}
\toprule
 \textbf{Method} &
 \textbf{\#Param.} 
 & \textbf{PSNR} $\uparrow$ & \textbf{rFID} $\downarrow$
 \\
\midrule
 Vanilla DINO-AE & 125M & 
 19.44 & 1.84 \\ 
Vanilla DINO-AE & 268M & 
 19.56 & 1.79 \\
 SD-VAE & \textless 100M & \underline{24.44} & \underline{0.87} \\ 
 \rowcolor{cellgreen!10}\method{}-AE & 125M & $\mathbf{25.07}$ & $\mathbf{0.28}$ \\ 
\bottomrule
\end{tabular}
}
\end{minipage}
\hfill
\begin{minipage}{.5\textwidth}
 \begin{center}
     \centerline{\includegraphics[width=0.97\textwidth,height=2.5cm]{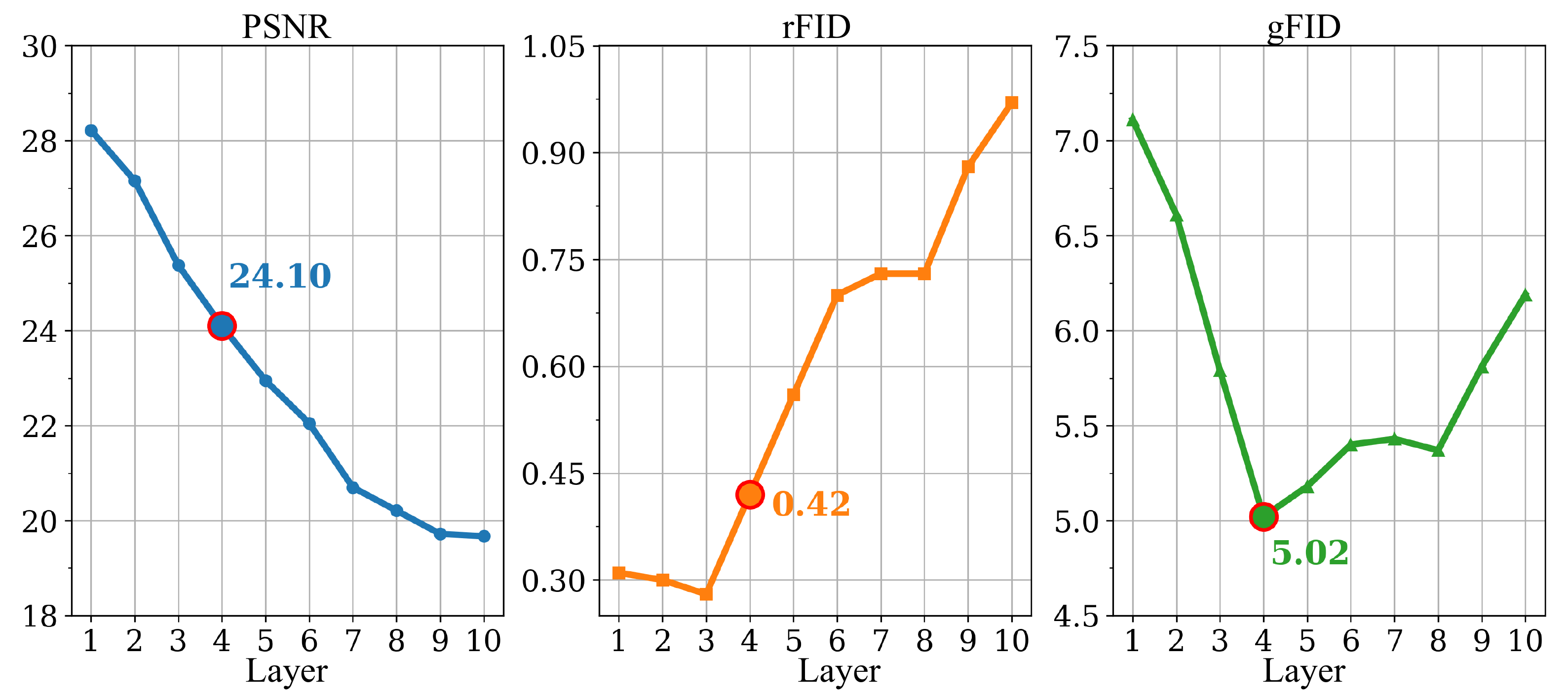}}
     \end{center}
\end{minipage}
\vspace{-.5cm}
\caption{
\textbf{Left:} ImageNet-1k $256\times256$ reconstructions with varying decoder sizes. Scaling the decoder size yields only marginal improvement. \textbf{Right:} Effect of shallow-layer choice on reconstruction and generation, and layer $4$ offers the best trade-off.}
\vspace{-.5cm}
\label{fig:ablation_layers}
\end{figure*}

\noindent\textbf{Shallow-layer branch ablation.}
We study how the choice of the shallow-layer feature affects the texture–semantic trade-off by evaluating reconstruction and generation (40 epochs) results. As shown in Fig.~\ref{fig:ablation_layers} (right) and Fig.~\ref{fig:dino_layers}, earlier layers (e.g., layer 1–3) retain stronger low-level structural signals and thus achieve better rFID and PSNR, but their generative performance (gFID) is weaker. As the shallow branch moves deeper, semantics become stronger and gFID improves, while reconstruction quality consistently degrades. Layer 4 yields the best overall balance, achieving the lowest gFID with competitive rFID and PSNR, indicating that mid-level features provide a good compromise between texture preservation and semantic alignment. We therefore adopt \textbf{layer 4} as the default shallow branch in both \method{}-AE and \method{}-VQ.

\noindent\textbf{Dominant-Subspace Quantization.}
We further analyze Dominant-Subspace Quantization (DSQ) by varying the semantic subspace dimensionality and the way components are selected, as summarized in Tab.~\ref{tab:ab_pca_quant}. Using all $768$ channels for VQ leads to severe codebook collapse (only $2.3\%$ usage) and semantic replacement (see Fig.~\ref{fig:dinov2_dino_rec}-(B)), which is a clear sign of quantization optimization failure under frozen high-dimensional semantic features. Restricting quantization to the top principal components both stabilizes training and improves performance. Top-64 dimensions give the best reconstruction ($1.44$ rFID, $20.65$ PSNR), while top-32 dimensions achieve the best generative quality ($7.25$ gFID) with only a minor drop in reconstruction. In contrast, randomly selecting 32 dimensions degrades both rFID and gFID, and quantizing the bottom 32 dimensions causes dramatic failure (gFID $\approx 216.9$), confirming that low-variance channels contribute little semantic structure and are largely noisy. Based on this trade-off, we use the \textbf{top-32 dominant dimensions} as the default semantic subspace in DSQ. Overall, these results support our hypothesis that high-dimensional DINO features exhibit an uneven variance distribution, and that selective quantization over principal components is essential for stable, semantics-preserving VQ.

\begin{table}[t]
\centering
\caption{
\textbf{Ablation on Dominant-Subspace Quantization.} Quantizing top principal dimensions improves reconstruction and generation, while larger subspaces reduce codebook usage and full $768$ dimension quantization collapses.}
\vspace{-.1cm}
\label{tab:ab_pca_quant}
\resizebox{0.9\linewidth}{!}{
\setlength{\tabcolsep}{4.5mm}
\begin{tabular}{lccccc}
\toprule
 \textbf{Type} &\textbf{\#Dim} 
 & \textbf{Usage} $\uparrow$ 
 & \textbf{rFID} $\downarrow$ & \textbf{PSNR} $\uparrow$ & \textbf{gFID}$\downarrow$ \\
\midrule
Top &768& 2.3\% &-- & --&-- \\
Top &64& 99.9\% &$\mathbf{1.44}$ & $\mathbf{20.65}$&\underline{8.24} \\
\rowcolor{cellgreen!10}
Top &32& 100.0\% &\underline{1.56} & \underline{20.64}&$\mathbf{7.25}$ \\
Random &32& 100.0\%& 2.19&20.16&11.30\\
Bottom&32&99.9\%&6.77&20.42&216.9\\ 
\bottomrule
\end{tabular}
}
\vspace{-.5cm}
\end{table}

\noindent\textbf{Decoder Size.}

To disentangle representation limitations from decoder capacity, we vary the decoder size while keeping the DINO-base encoder frozen and feeding only the high-level feature $\mathbf{F}_{\text{sem}}$ into the decoder which we refer to as \textbf{vanilla DINO-AE}. As shown in Fig.~\ref{fig:ablation_layers} (left), scaling the vanilla DINO-AE decoder from 125M to 268M parameters brings only marginal gains (PSNR from $19.44$ to $19.56$, rFID from $1.84$ to $1.79$), and both variants remain far behind SD-VAE, whose decoder is even smaller (\textless100M, $24.44$ PSNR, $0.87$ rFID). In contrast, with the same 125M decoder budget, \method{}-AE attains $25.07$ PSNR and $0.28$ rFID by using the information-complete latent. This indicates that the main bottleneck lies in the information content of the latent itself, where high-level DINO features alone lack sufficient high-frequency structure for faithful reconstruction, and simply increasing decoder capacity cannot compensate for this deficiency. The fused multi-layer dual-branch design in \method{} is therefore crucial for unlocking the full potential of frozen DINO representations.
\section{Conclusion}
In this work, we revisit visual tokenization design through the lens of frozen VFMs and identify two key challenges: the texture–semantic trade-off in reconstruction and the instability of quantizing frozen high-dimensional semantic features. To address these challenges, we introduce \method{}, a unified tokenizer built on DINO that constructs an information-complete latent for continuous tokenization and introduces Dominant-Subspace Quantization (DSQ) to stably discretize principal semantic components. Experiments on ImageNet-1k $256\times256$ show that \method{} achieves state-of-the-art reconstruction quality and strong continuous- and discrete-space class-conditional generation, demonstrating that frozen VFMs can be directly adapted into high-fidelity, semantically aligned visual tokenizers for next-generation latent generative models.
\clearpage
\onecolumn
\setcounter{page}{1}
\begin{center}
    {\LARGE\bfseries Appendix}
\end{center}
\vspace{1em}
\appendix
\section{Ablation on Semantic Similarity Loss}
\label{sec:sup_simloss}
We provide an ablation study on the proposed semantic similarity loss $\mathcal{L}_{sem}$ for \method{}-VQ.
The goal of $\mathcal{L}_{sem}$ is to regularize token alignment with the frozen DINO semantics, enhancing semantic structure under DSQ.. As shown in Table~\ref{tab:simloss_ablation}, introducing $\mathcal{L}_{sem}$ consistently improves both reconstruction and generation.
In particular, the rFID is reduced from $1.29$ to $1.09$, while the downstream generation quality is also improved, with gFID decreasing from $4.71$ to $3.85$ and IS increasing from $212.71$ to $237.12$. 
These results indicate that semantic alignment not only benefits representation, but also leads to a more generation-friendly latent space.

\begin{figure}[ht]
\begin{center}
\centerline{\includegraphics[width=0.85\textwidth]{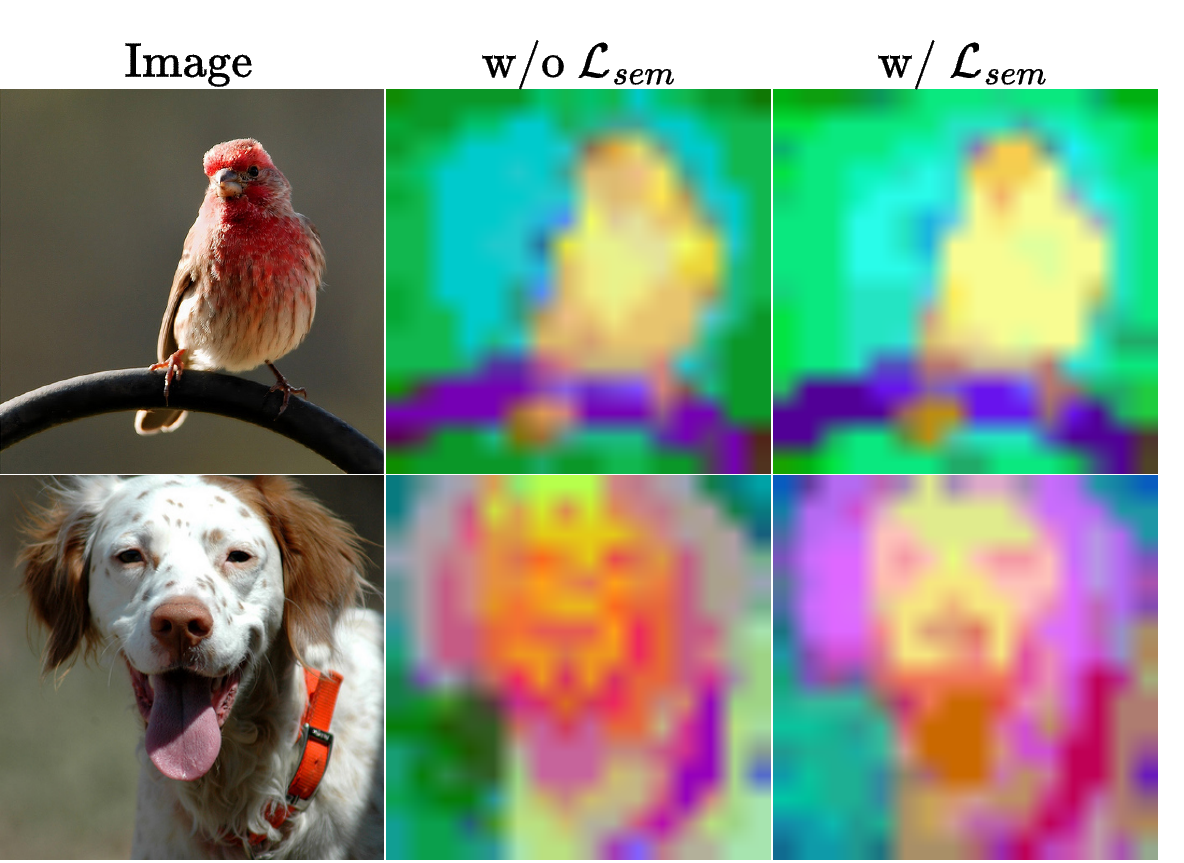}}
\caption{
\textbf{PCA visualization of latent features with and without the semantic similarity loss $\mathcal{L}_{sem}$.}
Introducing $\mathcal{L}_{sem}$ leads to a more organized latent space, with more compact semantic structures and less noise.
}
\label{fig:pca_sem}
\end{center}
% \vspace{-20pt}
\end{figure}

Figure~\ref{fig:pca_sem} further visualizes the effect of $\mathcal{L}_{sem}$ by applying PCA to the latent features. Without $\mathcal{L}_{sem}$, the latent embeddings already retain coarse semantic structure due to DSQ, but remain noisier and less organized. After introducing $\mathcal{L}_{sem}$, the embeddings exhibit more compact semantic clusters and clearer structures with reduced noise. This observation is consistent with the quantitative improvements in Table~\ref{tab:simloss_ablation}, and suggests that $\mathcal{L}_{sem}$ helps preserve more accurate semantic information in the learned tokenizer.

\begin{table}[h]
\centering
\caption{\textbf{Ablation of the semantic similarity loss $\mathcal{L}_{sem}$ on \method{}-VQ for ImageNet-1k 256$\times$256.}}
\label{tab:simloss_ablation}
\resizebox{0.85\linewidth}{!}{
\setlength{\tabcolsep}{5pt}
\begin{tabular}{lccc}
\toprule
\textbf{Methods} 
& \textbf{rFID}$\downarrow$ 
& \textbf{gFID}$\downarrow$ 
& \textbf{IS}$\uparrow$ \\
\midrule
\method{}-VQ w/o $\mathcal{L}_{sem}$ & 1.29 & 4.71 & 212.71 \\
\method{}-VQ w/ $\mathcal{L}_{sem}$  & \textbf{1.09} & \textbf{3.85} & \textbf{237.12} \\
\bottomrule
\end{tabular}
}
\end{table}

\section{Downstream Linear Probing Evaluation}
\label{sec:sup_linearprobe}

We further evaluate the semantic quality of different latent representations using linear probing on ImageNet-1k. For all compared methods, the backbone or tokenizer is frozen and only a linear classifier is trained on top of the extracted representation. We follow the standard DINO linear-probing protocol and report the corresponding latent layout for each method ($256$ tokens for 1D tokenizers and $16 \times 16$ for spatial tokenizers). This evaluation complements reconstruction and generation metrics by measuring how much category-level semantic information is preserved in the learned latent space.

The results are summarized in Table~\ref{tab:linear_prob_supp}. Among visual foundation models (VFMs), DINOv3-b achieves the strongest reference performance with $84.9\%$ top-1 accuracy, slightly outperforming DINOv2-b. For discrete tokenizers, \method{}-VQ achieves $77.2\%$ top-1 accuracy, substantially surpassing VFMTok~\cite{zheng2025vfmtok}, indicating that the proposed DSQ and $\mathcal{L}_{sem}$ preserve stronger dominant semantics. For continuous tokenizers, \method{}-AE reaches $85.0\%$ top-1 accuracy, which is slightly higher than DINOv3-b and also stronger than RAE~\cite{zheng2025rae} and SVG~\cite{shi2025svg} in our comparison. We additionally report reconstruction rFID as a reference to show the semantics-texture trade-off. In particular, RAE explicitly uses a frozen pretrained encoder and directly inherits the representation quality of the underlying encoder, while its DINOv2-b-based setting reports $84.5\%$ top-1 accuracy and $0.49$ rFID. Compared with this reference, \method{}-AE retains the full semantic representation from the DINO backbone, and further enriches it with an additional shallow-layer branch, achieving a substantially lower rFID of $0.28$, suggesting a more favorable balance between semantic preservation and reconstruction fidelity.

\begin{table}[h]
\centering
\caption{\textbf{Linear probing and reconstruction results on ImageNet-1k.}}
\label{tab:linear_prob_supp}
\resizebox{0.78\linewidth}{!}{
\setlength{\tabcolsep}{5.5pt}
\begin{tabular}{lcccc}
\toprule
 \textbf{Type} &
 \textbf{Method} &
 \textbf{Latent} &
 \textbf{Acc-Top1}$\uparrow$ &
 \textbf{rFID}$\downarrow$ \\
\midrule
\multirow{4}{*}{VFM}
& SigLIP-b~\cite{tschannen2025siglip} & $16\times16$ & 79.1 & -- \\
& MAE-b~\cite{he2022mae} & $16\times16$ & 68.0 & -- \\
& DINOv2-b~\cite{oquab2023dinov2} & $16\times16$ & 84.5 & -- \\
& DINOv3-b~\cite{simeoni2025dinov3} & $16\times16$ & 84.9 & -- \\
\midrule
% \multirow{2}{*}{VQ}
% & VFMTok~\cite{zheng2025vfmtok} & 256 & 69.4 & 1.13 \\
% & Ours-VQ & $16\times16$ & \textbf{77.2} & \textbf{1.10} \\
% \midrule
\multirow{4}{*}{AE}
& SD-VAE~\cite{esser2024sd3} & $16\times16$ & $\sim$8 & 0.87 \\
& RAE~\cite{zheng2025rae} & $16\times16$ & 84.5 & 0.49 \\
& SVG~\cite{shi2025svg} & $16\times16$ & 79.8 & 0.65 \\
& Ours-AE & $16\times16$ & \textbf{85.0} & \textbf{0.28} \\
\bottomrule
\end{tabular}
}
\end{table}

\section{Detailed Reconstruction Results}
We report additional reconstruction metrics in Table~\ref{tab:imagenet256recsup}, including LPIPS, PSNR, and SSIM.
These metrics complement rFID by measuring perceptual similarity, pixel-level fidelity, and structural consistency, respectively.
Overall, \method{} demonstrates strong reconstruction quality under both the continuous and discrete settings.
For continuous tokenizers, \method{}-AE matches the best reported rFID ($0.28$) among the compared methods, while also maintaining competitive perceptual and pixel fidelity.
Compared with RAE~\cite{zheng2025rae} and SVG~\cite{shi2025svg}, our method achieves substantially stronger reconstruction quality on pixel-level metrics, indicating that the proposed latent design attains a favorable balance between semantic preservation and image fidelity. Compared with VA-VAE~\cite{yao2025vavae}, \method{}-AE achieves the same rFID while using a high-dimensional semantic latent space designed to preserve both semantics and reconstruction quality.
For discrete tokenizers, \method{}-VQ achieves the best rFID ($1.10$) among the listed methods, outperforming previous VQ-based and VFM-based tokenizers.
Although some baselines achieve slightly stronger LPIPS or PSNR, \method{}-VQ offers a favorable trade-off between semantic preservation, quantization capacity, and reconstruction fidelity, which is also reflected in its stronger downstream linear probing and generation performance.
\begin{table*}[t]
\centering
\caption{\textbf{Reconstruction performance on ImageNet-1k 256$\times$256.} ``ukn.'' indicates methods trained with extra datasets or data settings not fully specified in the original source. Despite being trained solely on ImageNet-1k, \method{} achieves strong overall performance across multiple reconstruction metrics.}

\label{tab:imagenet256recsup}
% \vspace{-3pt}
\setlength{\tabcolsep}{1pt}
\resizebox{\linewidth}{!}
{
\begin{tabular}{llcccccc}
\toprule
\textbf{Type} & \textbf{Method} & \textbf{Dim} & \textbf{Size} & \textbf{rFID} $\downarrow$ & \textbf{LPIPS} $\downarrow$ & \textbf{PSNR} $\uparrow$ & \textbf{SSIM} $\uparrow$ \\
\midrule
\multirow{5}{*}{Continuous}
& SD-VAE\textsuperscript{ukn.}~\cite{rombach2022ldm} & 16 & - & 0.87 & 0.1363 & 24.44 & 0.698 \\
% & Hunyuan-VAE\textsuperscript{ukn.}~\cite{kong2024hunyuanvideo} & 16 & - & 1.58 & \underline{0.0737} & \underline{29.57} & 0.849 \\
% & Wan-VAE\textsuperscript{ukn.}~\cite{wan2025} & 16 & - & 0.76 & \textbf{0.0516} & \textbf{29.63} & \textbf{0.863} \\
& SVG~\cite{shi2025svg} & 392 & - & 0.65 & 0.1900 & 23.89 & 0.650 \\
& RAE~\cite{zheng2025rae} & 768 & - & {0.49} & - & 19.23 & 0.620 \\
& VA-VAE~\cite{yao2025vavae} & 32 & - &  {0.28} & {0.0962} & {27.96} & {0.790} \\
& Ours-AE & 832 & - &  {0.28} & {0.0995} & {25.07} & {0.740} \\
\midrule
\multirow{8}{*}{Discrete}
& VQGAN~\cite{van2017vqvae} & 256 & 16384  & 4.98 & 0.2843 & 20.00 & 0.629 \\
& LlamaGen~\cite{sun2024llamagen} & 8 & 16384  & 2.19 & 0.2281 & 20.79 & 0.675 \\
& TiTok~\cite{yu2024titok} & 16 & 4096 & 1.66 & - & 20.01 & - \\
& Open-MAGVIT2~\cite{luo2024oplfq} & 18 & 262144 & 1.17 & 0.2038 & 21.90 & - \\
& VAR~\cite{tian2025var} & 32 & 4096 & - & - & 21.30 & 0.647  \\
& VFMTok~\cite{zheng2025vfmtok} & 12 & 16384  & {1.13} & 0.2680 & 19.91 & 0.488 \\
& IBQ~\cite{shi2025ibq} & 256 & 16384  & {1.37} & 0.2235 & - & - \\
& Ours-VQ & 832& $16384 \times 2$ & {1.10} & {0.2382} & {20.45} & {0.544} \\
\bottomrule
\end{tabular}
}
\end{table*}

\section{Further Analysis of PCA on DINO Features}
\label{sec:sup_pca}
We present additional PCA-based analyses to better understand the semantic structure of DINO features and to further justify the proposed DSQ strategy.
Our observations consistently show that the variance of DINO features is highly concentrated in a small subset of channels, while the remaining channels contribute substantially less.
This behavior motivates allocating more modeling or quantization capacity to the principal semantic components.

\begin{figure}[h]
\begin{center}
\centerline{\includegraphics[width=0.85\textwidth]{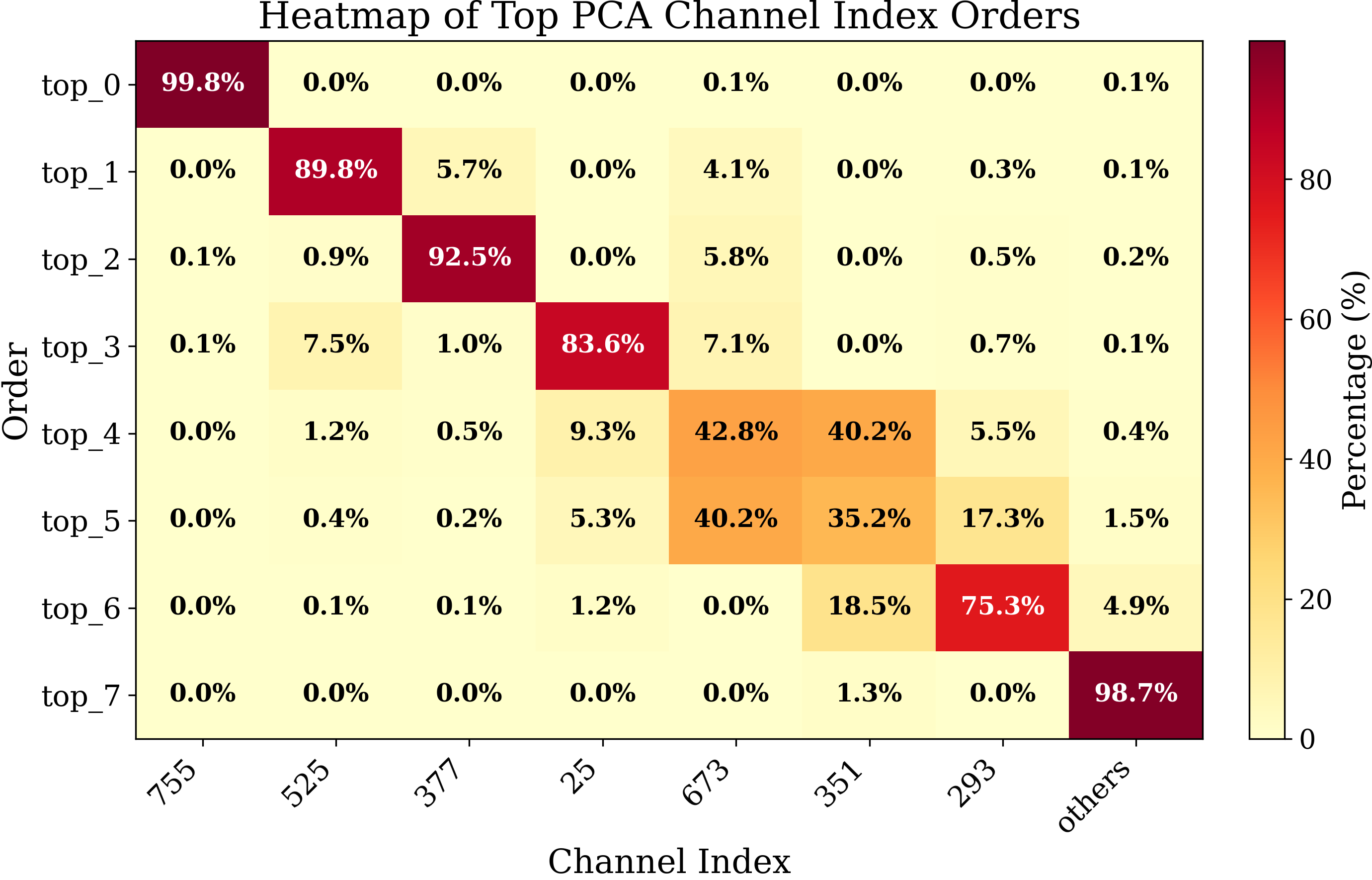}}
\caption{
\textbf{PCA on DINO features for all ``ruler'' images in ImageNet reveals globally consistent channel importance.}
The top channels are highly consistent across diverse instances, and the top\_0 channel (index 755) is activated in $99.8\%$ of all cases.
}
\label{fig:pca_map}
\end{center}
% \vspace{-20pt}
\end{figure}

\subsection{Global Consistency of PCA Top Channels}
To clarify the motivation of applying DSQ, we examine the global consistency of PCA components across intra-class variations. Specifically, we analyze all $1{,}300$ validation images of the class ``ruler'' in ImageNet-1k, which exhibit diverse colors, shapes, materials, and backgrounds. 
As shown in Figure~\ref{fig:pca_map}, the heatmap demonstrates a strong \textbf{global consistency} in channel importance across all instances, with a clear diagonal structure among the top channels. 
Notably, the top\_0 channel (index $755$) is activated in $99.8\%$ of all cases ($1298/1300$). 
This result indicates that the most dominant PCA channels are not random artifacts, but highly stable semantic directions shared across large appearance variations.
% Each image is resized and center-cropped to $512\times512$ resolution, and the top-8 PCA channel indices of each image are recorded.

\begin{figure}[ht]
\begin{center}
\centerline{\includegraphics[width=0.85\textwidth]{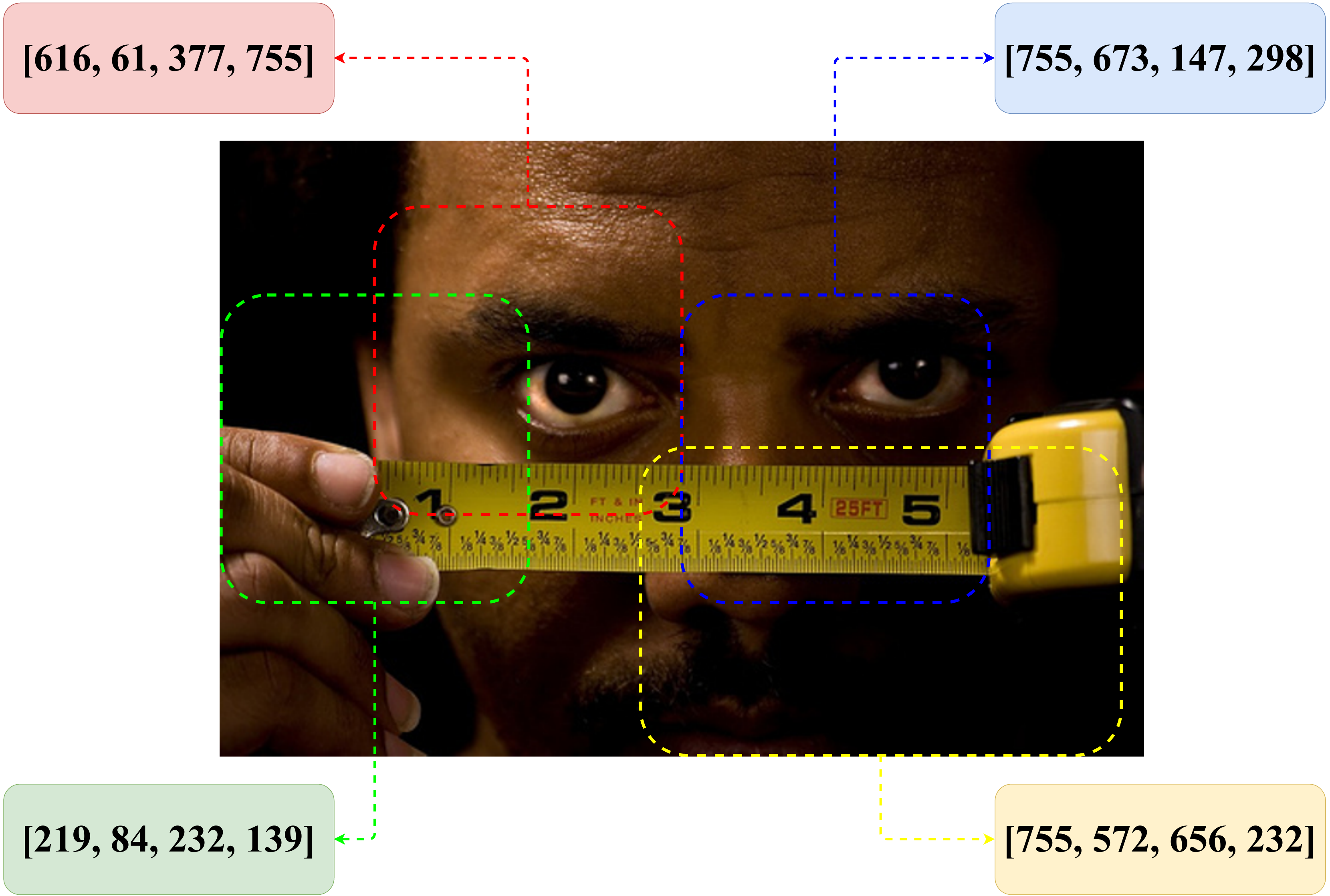}}
\caption{
\textbf{PCA of DINO embeddings under different image crops.}
Different crops induce different dominant channels, showing that the leading PCA components are sensitive to semantic content and local structures.
}
\label{fig:pca_crop}
\end{center}
% \vspace{-20pt}
\end{figure}

\subsection{Channel Shifts Across Different Image Crops}
We further investigate the sensitivity of PCA components to semantic changes using different crops from the same image. 
As shown in Figure~\ref{fig:pca_crop}, crops with distinct semantic emphasis lead to different dominant channels, highlighting the spatial sensitivity of DINO’s representation. 
The \textcolor{red}{red} crop, dominated by the eye region, shifts its top principal channel from $755$ to $616$, indicating a semantic reorientation toward fine-grained details. 
The \textcolor{green!70!black}{green} crop, which focuses on a local finger region, changes its top channel to $219$, reflecting localized texture dominance. 
By contrast, the \textcolor{yellow!80!black}{yellow} and \textcolor{blue}{blue} crops, which preserve the ruler and sufficient surrounding context, keep the original dominant channel ($755$). 
These observations demonstrate that PCA on DINO features is not only globally consistent, but also sensitive to different semantic contents.

\subsection{Long-Tail Distribution of PCA Eigenvalues}
\begin{figure}[h]
\begin{center}
\centerline{\includegraphics[width=0.85\textwidth]{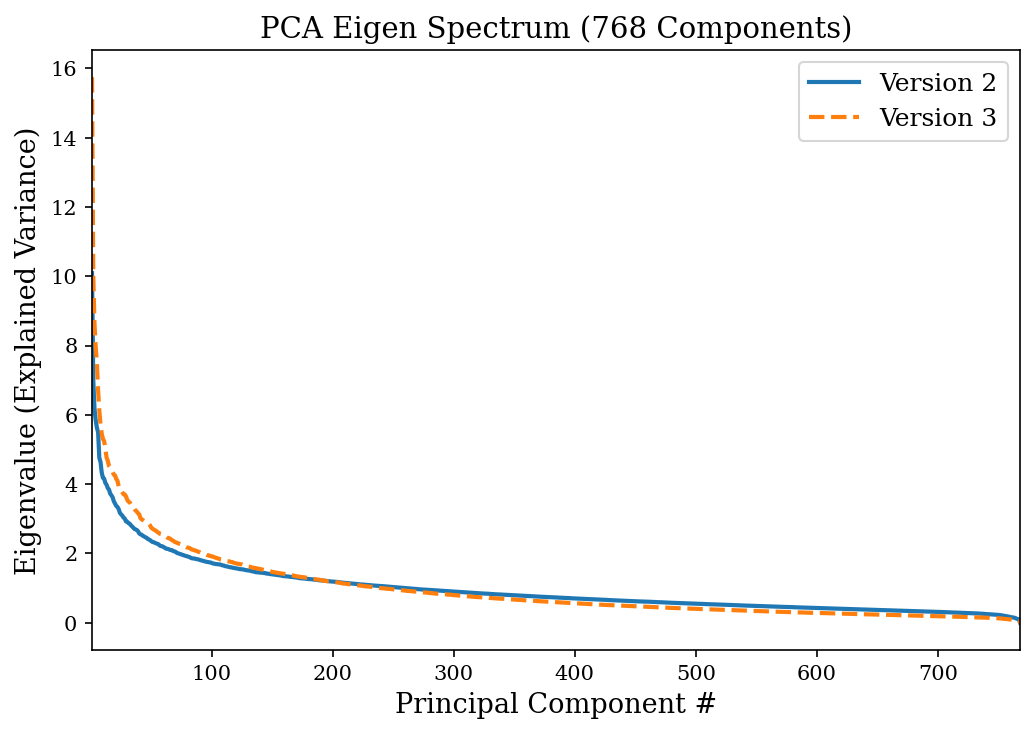}}
\caption{
\textbf{PCA eigenvalue distribution of DINO-v2 and DINO-v3 features.}
The long-tail spectrum indicates that only a few dominant channels capture most of the representational variance.
}
\label{fig:pca_dist}
\end{center}
% \vspace{-20pt}
\end{figure}

We analyze the eigenvalue spectrum of PCA applied to DINO features to understand how variance is distributed across channels. 
As illustrated in Figure~\ref{fig:pca_dist}, the eigenvalue distribution follows a clear long-tail pattern, indicating that only a small number of principal components account for the majority of the representational variance. 
This concentration suggests that semantic information is unevenly distributed in the feature space, providing direct motivation for extracting dominant subspaces during quantization.

\subsection{Complete DINO Sorted Channels Visualization}
\begin{figure*}[tb]
\begin{center}
\centerline{\includegraphics[width=0.97\textwidth]{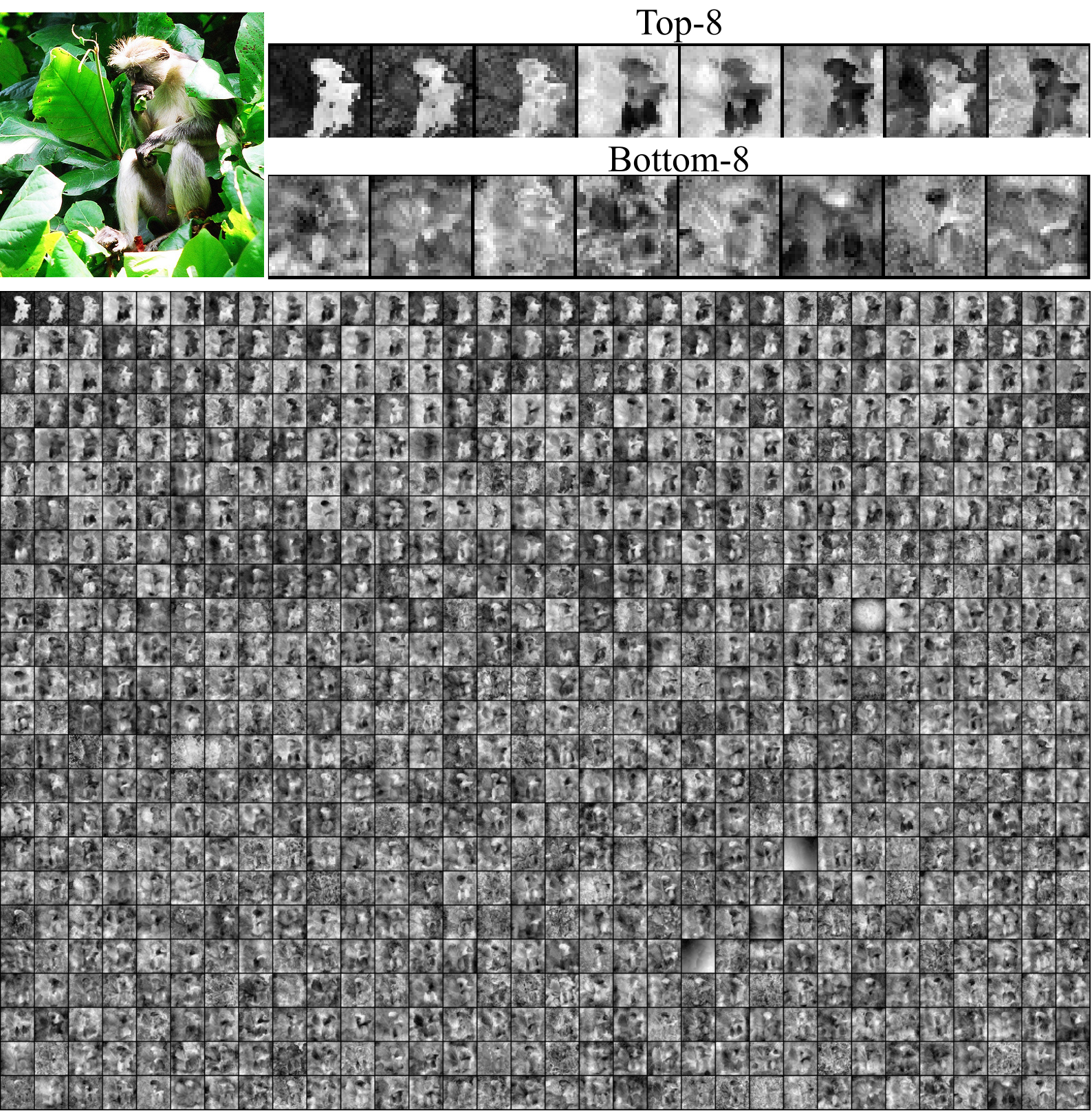}}
\caption{
\textbf{Visualization of all PCA-sorted $768$ channels of DINO-v3-base features.}
Channels are ordered by PCA importance. Top channels exhibit clear semantic structure, while bottom channels appear increasingly noisy.
}
\label{fig:pca_channel_all}
\end{center}
\end{figure*}

\begin{figure*}[ht]
\begin{center}
\centerline{\includegraphics[width=0.97\textwidth]{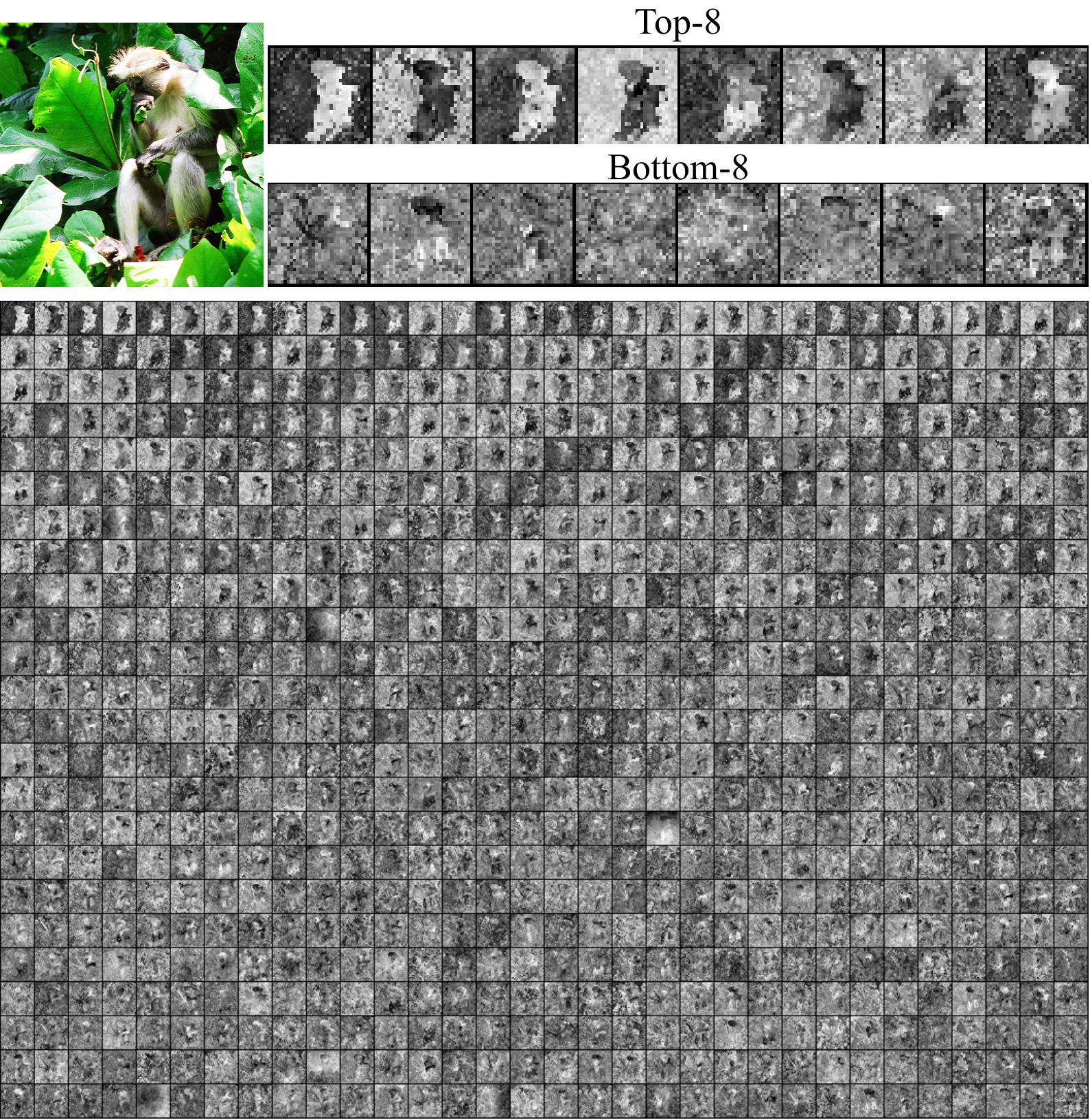}}
\caption{
\textbf{Visualization of all PCA-sorted $768$ channels of DINO-v2-base features.}
Similar to DINO-v3-base, the most dominant channels are much more interpretable than the low-ranked channels.
}
\label{fig:pca_channel_all_v2}
\end{center}
\end{figure*}

We provide a complete visualization of all $768$ PCA-sorted channels from DINO-v3-base and DINO-v2-base in Figure~\ref{fig:pca_channel_all} and Figure~\ref{fig:pca_channel_all_v2}. 
The top channels show clear semantic structures and object-aligned patterns, while the bottom channels become increasingly noisy and difficult to interpret.
This progressive transition from structured to noisy channels further supports the dominant-subspace assumption used in our quantization design.

\section{Theoretical Analysis of the Distance Concentration Phenomenon}
\label{sec:sup_distance}
We provide a brief theoretical background on why Euclidean ($L_2$) distance–based lookup in traditional vector quantization becomes unreliable in high-dimensional spaces. 
As shown in prior work~\cite{beyer1999nearest,aggarwal2001distance,peng2023interpreting}, the distance concentration behavior of Minkowski distances is a direct consequence of concentration-of-measure phenomena in high dimensions.

Consider a set of $n$ data points $\mathcal{P}=\{\mathbf{P}_1^d,\ldots,\mathbf{P}_n^d\}$, where each point $\mathbf{P}_i^d = (p_i^1,\ldots,p_i^d) \in \mathbb{R}^d$, and a query point $\mathbf{Q}^d = (q^1,\ldots,q^d) \in \mathbb{R}^d$. The $p$-norm (Minkowski) distance between $\mathbf{P}_i^d$ and $\mathbf{Q}^d$ is defined as
\begin{equation}
    L_p(\mathbf{P}_i^d,\mathbf{Q}^d)
    = \left( \sum_{k=1}^{d} \big|p_i^k - q^k\big|^p \right)^{\!1/p},
\end{equation}
which reduces to the standard Euclidean distance when $p=2$.

Formally, as the dimensionality $d$ grows, the relative contrast between the farthest and nearest samples vanishes:
\begin{equation}
    \lim_{d \to \infty} 
    \frac{D^d_{\max} - D^d_{\min}}{D^d_{\min}} = 0,
\end{equation}
where
\begin{equation}
\begin{aligned}
D^d_{\max} &= \max_{i=1,\ldots,n} \big\|\mathbf{P}_i^d - \mathbf{Q}^d\big\|_p,\\
D^d_{\min} &= \min_{i=1,\ldots,n} \big\|\mathbf{P}_i^d - \mathbf{Q}^d\big\|_p.
\end{aligned}
\end{equation}
In other words, nearest and farthest neighbors become increasingly indistinguishable in high dimensions.
For high-dimensional visual tokens, this phenomenon weakens the effectiveness of plain $L_2$ lookup and motivates our DSQ strategy, which emphasizes dominant semantic channels rather than treating all channels uniformly.

\section{More Qualitative Visualizations}

\begin{figure*}[htbp]
\begin{center}
\centerline{\includegraphics[width=0.98\textwidth]{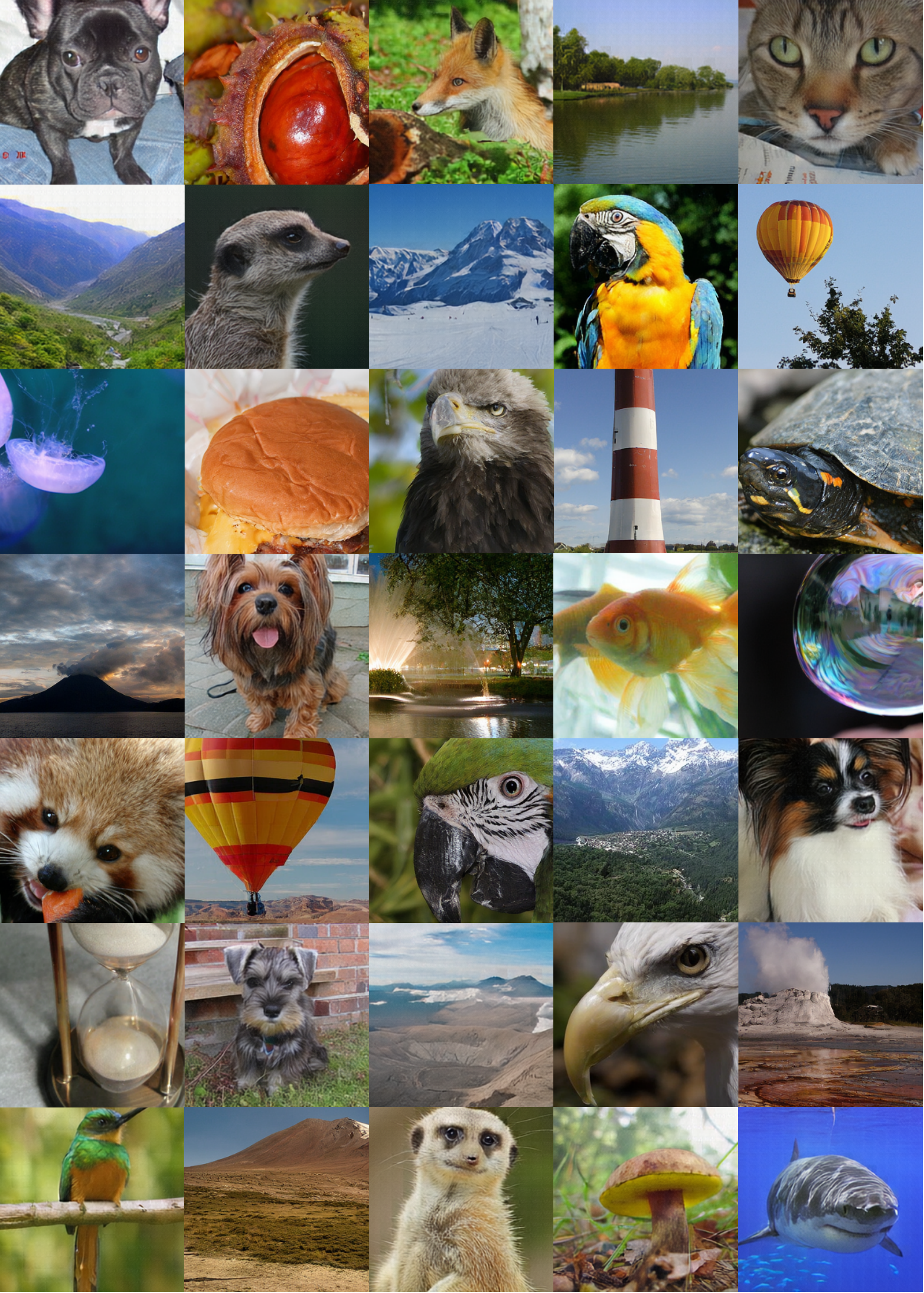}}
\caption{
\textbf{More qualitative image generation results of \method{}-AE.}
\method{}-AE produces high-quality and diverse class-conditional generations across different semantic categories.
}
\label{fig:gen_vis}
\end{center}
% \vspace{-20pt}
\end{figure*}

\begin{figure*}[htbp]
\begin{center}
\centerline{\includegraphics[width=0.98\textwidth]{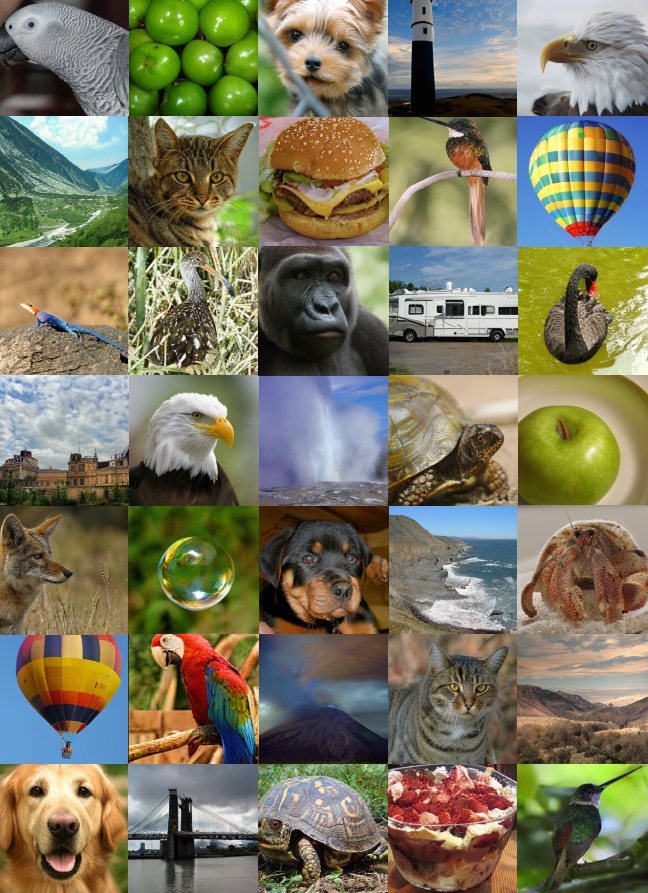}}
\caption{
\textbf{More qualitative image generation results of \method{}-VQ.}
\method{}-VQ produces high-quality and diverse class-conditional generations across different semantic categories.
}
\label{fig:gen_vis_vq}
\end{center}
% \vspace{-20pt}
\end{figure*}

\begin{figure*}[t]
\begin{center}
\centerline{\includegraphics[width=0.94\textwidth]{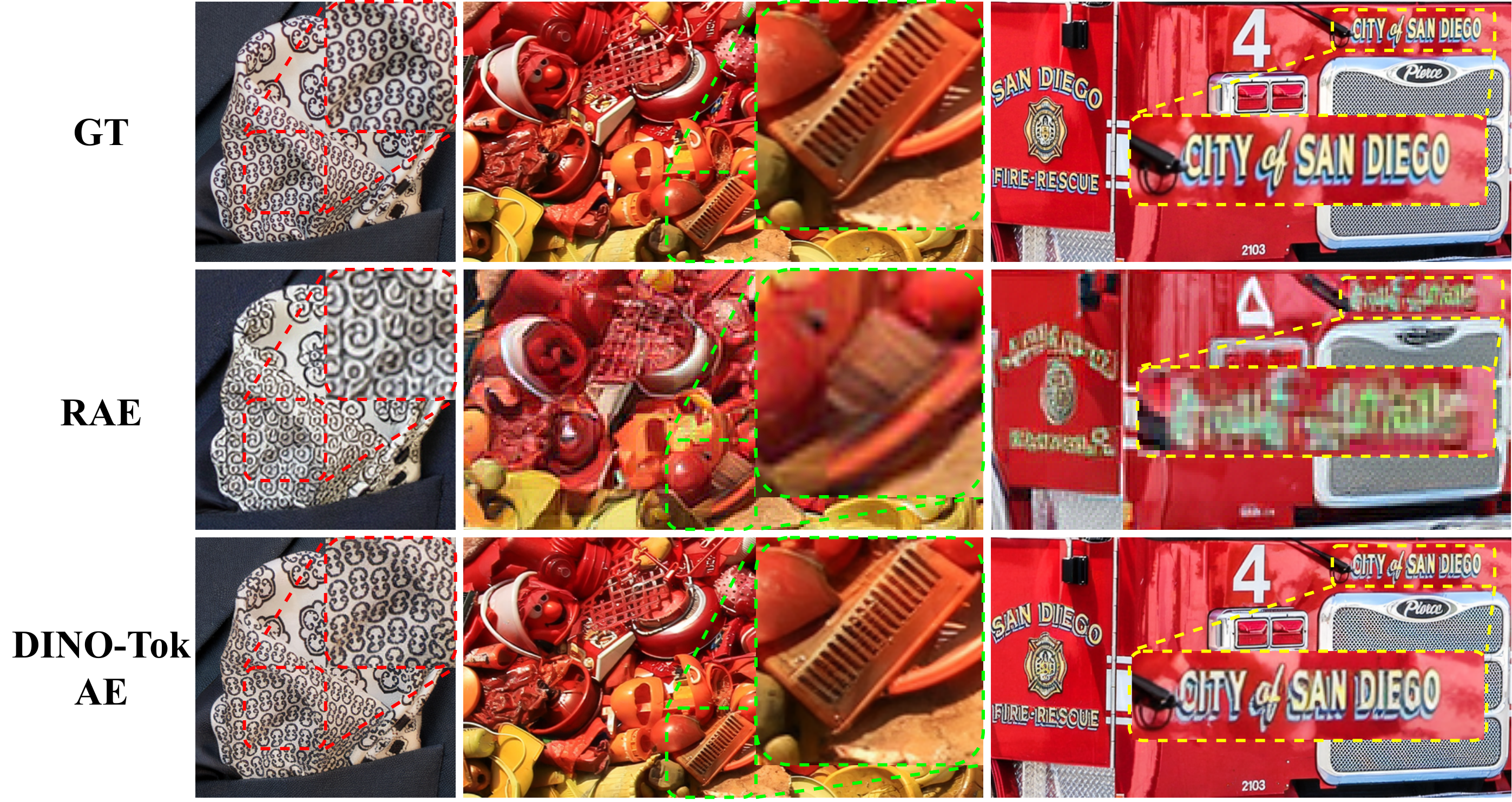}}
\caption{
% PCA on DINO features for all "ruler" images in ImageNet reveals a strong global consistency in channel importance. This is evidenced by the clear diagonal pattern across the top-8 channels, with the top-1 channel (\#755) activating for 99.8\% of all cases.
\textbf{More qualitative results of continuous tokenizer reconstruction.} \method{} can reconstruct more faithful details and higher-fidelity images compared to the baseline.
}
\label{fig:recon_vis}
\end{center}
\end{figure*}

\begin{figure*}[ht]
\begin{center}
\centerline{\includegraphics[width=0.91\textwidth]{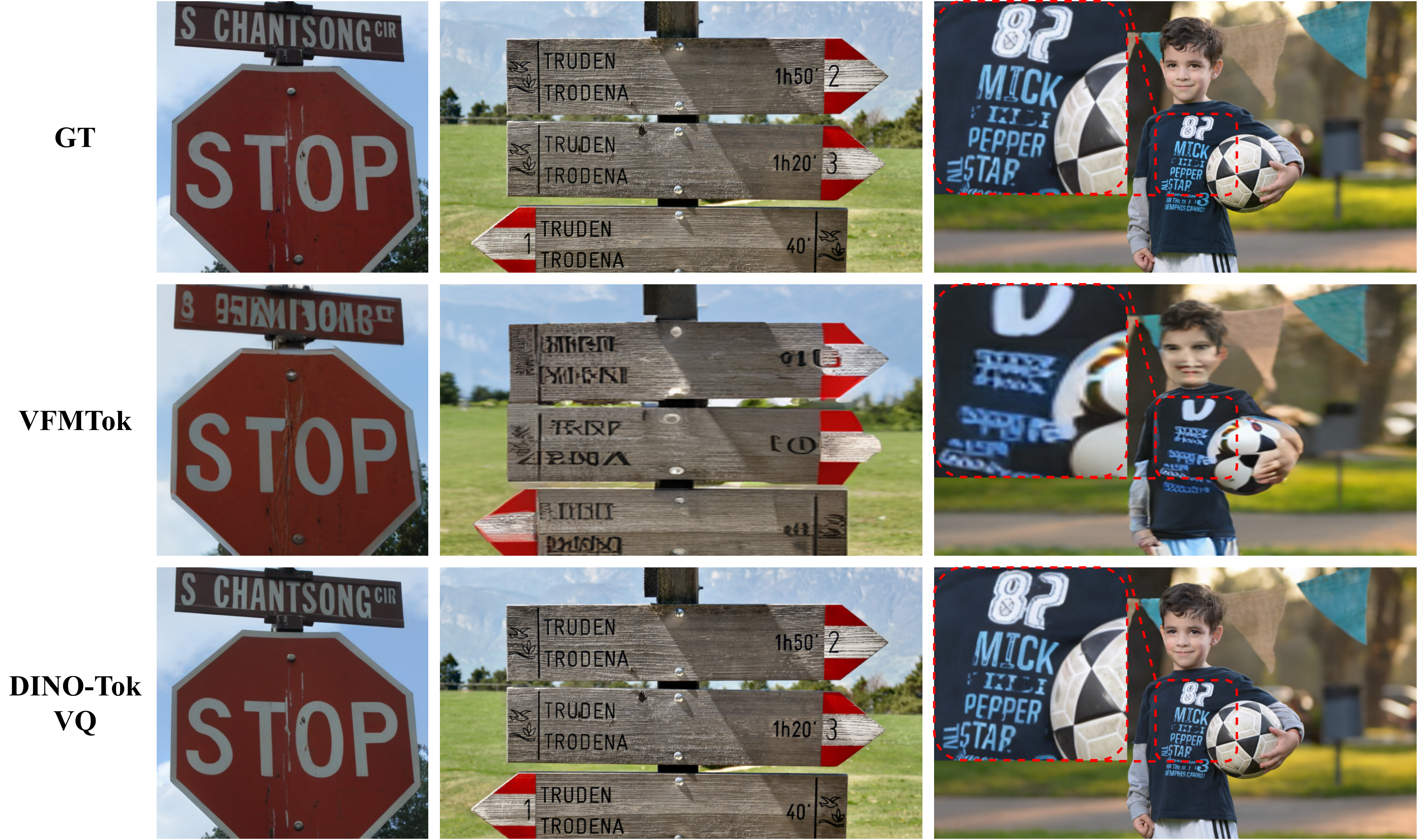}}
\caption{
% PCA on DINO features for all "ruler" images in ImageNet reveals a strong global consistency in channel importance. This is evidenced by the clear diagonal pattern across the top-8 channels, with the top-1 channel (\#755) activating for 99.8\% of all cases.
\textbf{More qualitative results of discrete tokenizer reconstruction.} \method{} can reconstruct more faithful details and higher-fidelity images compared to the baseline.
}
\label{fig:recon_visvq}
\end{center}
\end{figure*}

We provide additional qualitative visualizations of generation and reconstruction in Figure~\ref{fig:gen_vis} and Figure~\ref{fig:gen_vis_vq}. \method{}-AE and \method{}-VQ produces diverse and visually plausible class-conditional samples with fine local details.
Figures~\ref{fig:recon_vis} and~\ref{fig:recon_visvq} further show that both \method{}-AE and \method{}-VQ reconstruct faithful image structures, object boundaries, and fine textures.

\clearpage

\bibliographystyle{splncs04}
\bibliography{main}
\end{document}